\documentclass{article}

\usepackage{microtype}
\usepackage{graphicx}
\usepackage{subfigure}
\usepackage{booktabs} 
\usepackage{enumitem}

\usepackage{hyperref}

\usepackage[accepted]{icml2024}

\usepackage{amsmath}
\usepackage{amssymb}
\usepackage{amsfonts}
\usepackage{bbm}
\usepackage{mathtools}
\usepackage{amsthm}
\usepackage{caption}
\usepackage{float}
\usepackage{booktabs}

\usepackage{algorithm}
\usepackage{algorithmic}

\usepackage[capitalize,noabbrev]{cleveref}

\newif\iffinal
\finaltrue

\usepackage{xcolor}

\iffinal
    \newcommand{\tianhao}[1]{}
    \newcommand{\ruoxi}[1]{}
    \newcommand{\yc}[1]{}
    \newcommand{\add}[1]{#1}
    \newcommand{\addcheck}[1]{#1}
    \newcommand{\james}[1]{}
    \newcommand{\tianji}[1]{}
\else
    \newcommand{\tianhao}[1]{{\bf \textcolor{purple}{[Tianhao: #1]}}}
    \newcommand{\ruoxi}[1]{{\bf \textcolor{violet}{[Ruoxi:#1]}}}
    \newcommand{\add}[1]{\textcolor{red}{#1}}
    \newcommand{\yc}[1]{{\color{blue} [{\bf YC:} #1]}}
    \newcommand{\addcheck}[1]{\textcolor{teal}{#1}}
    \newcommand{\james}[1]{{\color{blue} {\bf James:} #1}}
    \newcommand{\tianji}[1]{{\bf \textcolor{orange}{[Tianji: #1]}}}
\fi

\usepackage{cleveref}
\usepackage{bbm}
\usepackage{xspace}

\newtheorem{theorem}{Theorem}

\newtheorem{definition}[theorem]{Definition}

\newtheorem{remark}{Remark}

\newtheorem{remark-star}{Remark}
\newtheorem{remark-star-1}{Remark}

\newtheorem*{proof-sketch}{Proof Sketch}
\usepackage{booktabs}

\DeclareMathOperator*{\argmin}{\arg\!\min}
\DeclareMathOperator*{\argmax}{\arg\!\max}

\newcommand{\E}{\mathbb{E}}
\newcommand{\Var}{\mathrm{Var}}
\newcommand{\R}{\mathbb{R}}

\newcommand{\M}{\mathcal{M}}

\newcommand{\ind}{\mathbbm{1}}

\newcommand{\A}{\mathcal{A}}

\newcommand{\U}{v}
\newcommand{\test}{\mathrm{val}}

\newcommand{\metric}{\texttt{ValAcc}}

\newcommand{\unif}{\mathrm{Unif}}

\newcommand{\xval}{ x^{(\test)} }
\newcommand{\yval}{ y^{(\test)} }

\usepackage{tikz}

\newcommand{\sign}{ \texttt{sign} }

\newcommand{\Ualter}{\widetilde{\U}}

\newcommand{\testfunc}{\mathbf{h}}

\newcommand{\dataset}{N}
\newcommand{\Sgood}{S_{\text{clean}}}
\newcommand{\Sbad}{S_{\text{bad}}}
\newcommand{\Hnull}{H_{(0)}}
\newcommand{\Ha}{H_{(a)}}

\newcommand{\optS}{S_{*, \U}}

\newcommand{\ShapleySet}{\widehat S_{\phi(\U)}}

\newcommand{\F}{\mathcal{F}}
\newcommand{\Spec}{\textbf{TrueNeg}}
\newcommand{\Sens}{\textbf{TruePos}}

\newcommand{\classNull}{\F^{(0)}_{S_1, S_2}}
\newcommand{\classAlter}{\F^{(a)}_{S_1, S_2}}

\newcommand{\zero}{\mathbf{0}}

\newcommand{\trainSample}{\mathcal{S}_{\text{train}}}

\newcommand{\mon}{\text{MTM}\xspace}

\newcommand{\disc}{\texttt{cor}_{\rho}}
\newcommand{\mR}{\mathcal{R}}
\newcommand{\normR}{\bar \mR_{\U}}

\newcommand{\sampleRho}{\rho\text{-corr}}

\newcommand{\correlation}{$\rho$-consistency index\xspace}

\usepackage[textsize=tiny]{todonotes}

\icmltitlerunning{Rethinking Data Shapley for Data Selection Tasks: Misleads and Merits}

\begin{document}

\twocolumn[
\icmltitle{Rethinking Data Shapley for Data Selection Tasks: Misleads and Merits}




\begin{icmlauthorlist}
\icmlauthor{Jiachen T. Wang}{princeton}
\icmlauthor{Tianji Yang}{est}
\icmlauthor{James Zou}{stf}
\icmlauthor{Yongchan Kwon}{columbia}
\icmlauthor{Ruoxi Jia}{vt}
\end{icmlauthorlist}

\icmlaffiliation{princeton}{Princeton University}
\icmlaffiliation{vt}{Virginia Tech}
\icmlaffiliation{stf}{Stanford University}
\icmlaffiliation{columbia}{Columbia University}
\icmlaffiliation{est}{East China Normal University}

\icmlcorrespondingauthor{Jiachen T. Wang}{tianhaowang@princeton.edu}
\icmlcorrespondingauthor{Ruoxi Jia}{ruoxijia@vt.edu}

\icmlkeywords{Machine Learning, ICML}

\vskip 0.3in
]



\printAffiliationsAndNotice{} 

\begin{abstract}
Data Shapley provides a principled approach to data valuation and plays a crucial role in data-centric machine learning (ML) research. 
Data selection is considered a standard application of Data Shapley. 
However, its data selection performance has shown to be inconsistent across settings in the literature. 
This study aims to deepen our understanding of this phenomenon. 
We introduce a hypothesis testing framework and show that Data Shapley's performance can be no better than random selection without specific constraints on utility functions. 
We identify a class of utility functions, monotonically transformed modular functions, within which Data Shapley optimally selects data. 
Based on this insight, we propose a heuristic for predicting Data Shapley’s effectiveness in data selection tasks. 
Our experiments corroborate these findings, adding new insights into when Data Shapley may or may not succeed.

\end{abstract}

\section{Introduction}
\label{sec:intro}

\textbf{Data valuation and Data Shapley.} 
Data is the backbone of machine learning (ML) models, but not all data is created equally. In real-world scenarios, data often carries noise and bias, sourced from diverse origins and labeling processes \citep{northcutt2021pervasive}. Against this backdrop, \emph{data valuation} emerges as a growing research field, aiming to quantify the contribution of individual data sources for ML model training. 
Drawing on cooperative game theory, the use of the \emph{Shapley value} for data valuation was pioneered by \citep{ghorbani2019data,jia2019towards}. The Shapley value is a renowned solution concept in game theory for fair profit attribution \citep{shapley1953value}. In the context of data valuation, individual data points or sources are regarded as ``players'' in a cooperative game, and \emph{Data Shapley} refers to the data valuation techniques that use the Shapley value as the contribution measure for each data owner. 
As the \emph{unique} value notion that satisfies a set of axioms \cite{shapley1953value}, Data Shapley has rapidly gained popularity since its introduction in 2019 and is increasingly being recognized as a standard tool for evaluating data quality, particularly in critical domains like healthcare \citep{tang2021data, pandl2021trustworthy, bloch2021data, zheng2023exploiting}.



\textbf{Data selection: a standard application of Data Shapley.} 
Data selection is a natural and important application of Data Shapley that is frequently mentioned in the literature. Data selection involves choosing the optimal training set from available data sources to maximize final model performance. 
Given that Data Shapley is a principled measure of data quality, a natural approach is to prioritize data sources with the highest Data Shapley scores. Consequently, a common practice in the literature is choosing the subsets of data points with top Shapley value scores. 

\textbf{Motivation.} 
Empirical evidence regarding Data Shapley's effectiveness in data selection, however, has been inconsistent. 
\add{Some studies report that Data Shapley significantly outperforms random selection baselines \cite{tang2021data, jiang2023opendataval}, while others find its performance is no better than the random baseline \cite{kwon2022beta, kwon2023data}. Such a phenomenon is also reproduced in our experiments (e.g., Figure \ref{fig:selection-main} in Section \ref{sec:eval}), where Data Shapley's performance varies over different kinds of training data.} 
Such inconsistency is not only a confusing phenomenon but also poses practical challenges. In critical sectors where data-driven decisions are crucial, relying on Data Shapley for data selection could lead to flawed decision-making. The existing data valuation literature, while rich in application and theory, reveals a notable missing aspect in understanding the efficacy of Data Shapley for data selection. There is an absence of theory to clarify and explain under what circumstances Data Shapley might mislead or benefit data selection. Our study aims to fill in this missing aspect, providing insights that could significantly influence the understanding of data valuation and its practical applications.

Our contributions are summarized as follows:

\textbf{A theoretical explanation for Data Shapley's limitations in data selection.} We introduce a novel hypothesis testing framework tailored to analyze Data Shapley's efficacy in data selection. Our findings reveal that in the absence of specific structural assumptions about utility functions, Data Shapley's performance in data selection tasks can be no better than that of random guessing. This stems from the non-injective nature of the Shapley value transformation; distinct utility functions can result in identical Shapley values. Hence, there's a significant challenge for reliably comparing dataset utilities based solely on Data Shapley scores in an information-theoretic sense.

\textbf{When does Data Shapley work well for data selection?} 
Our analysis demonstrates that Data Shapley excels in scenarios where utility functions adhere to specific structures shaped by the inherent characteristics of datasets or learning algorithms. One such example is the utility functions for any reasonable learning algorithm trained on heterogeneous datasets comprising a mix of high-quality and low-quality data. We characterize a broad class of utility functions, termed \emph{monotonically transformed modular functions} (\mon), within which Data Shapley proves to be optimal for data selection. This class comprises the utility functions of widely used learning algorithms, such as kernel methods.

\textbf{A heuristic for predicting Data Shapley’s optimality for data selection.} 
\add{Based on the insights from the scenarios where Data Shapley works well, we propose a heuristic for predicting the effectiveness of Data Shapley in data selection tasks.} 
This approach approximates the original utility function with a \mon function, and the fitting quality is used as an indicator of Data Shapley's potential efficacy. 
\add{We uncover a connection between the optimal \mon approximation quality and \emph{consistency index}, a concept that measures the correlation between the utilities of different datasets. 
This suggests that when the utilities of two similar datasets are highly correlated, 
the utility function can be approximated by a \mon with decent fitting quality.}
Our experimental results reveal a strong correlation between the effectiveness of Data Shapley in data selection and the fitting residual of the \mon approximation. This correlation is further tied to the \add{consistency index} of the utility functions, providing a deeper understanding of the factors influencing Data Shapley's performance.

Overall, this work offers comprehensive theoretical and practical insights into Data Shapley's effectiveness in data selection, which marks a step towards understanding the optimal usage of data valuation techniques.

\section{Background}
\label{sec:background}


\textbf{Set-up of data valuation.} 
Let $\dataset = \{1, \ldots, n\}$ denotes a training set of size $n$. The objective of data valuation is to assign a score to each training data point in a way that reflects their contribution or quality towards downstream ML tasks. These scores are called \emph{data values}. 

\textbf{Utility functions.} 
The cornerstone of Data Shapley and other game theory-based data valuation methods lies in the concept of the \emph{utility function}. 
It is a set function $\U: 2^\dataset \rightarrow \R$ that maps any subset of the training set $\dataset$ to a score indicating the usefulness of the data subset. $2^\dataset$ represents the power set of $\dataset$, i.e., the set of all subsets of $\dataset$, including the empty set and $\dataset$ itself. 
For classification tasks, a common choice for $\U$ is the validation accuracy of a model trained on the input subset. Formally, $\U(S) := \metric(\A(S))$, where $\A$ is a learning algorithm that takes a dataset $S$ as input and returns a model, and $\metric$ is a metric function used to assess the model's performance, e.g., the accuracy of a model on a hold-out validation set. 

\textbf{Notations \& assumptions.} 
We sometimes denote $S \cup i := S \cup \{i\}$ and $S \setminus i := S \setminus \{i\}$ for singleton $\{i\}$, where $i \in \dataset$ is a single data point from $\dataset$. We denote the data value of $i$ computed from $\U$ as $\phi_i(\U)$, and $\phi(\U) := (\phi_1(\U), \ldots, \phi_n(\U))$ the vector of data values for each $i \in \dataset$. We use $S \sim \unif(\dataset)$ to denote sampling a subset $S$ from $2^\dataset$ uniformly at random. When the context is clear, we write $\E_S$ or $\Var_S$ for expectation and variance taken over the randomness of $S$, while omitting the sampling distribution. 
Without loss of generality, throughout the paper, we assume $\U(S) \in [0, 1]$ and $\U(\varnothing) = 0$. We note that sometimes it is convenient to view $\U$ as a vector with $2^n-1$ entries, where each entry corresponds to $\U(S)$ for a non-empty $S$. 


\textbf{Data Shapley.} The Shapley value is arguably the most widely studied scheme for data valuation. At a high level, it appraises each point based on the (weighted) average utility change caused by adding the point into different subsets. 

\begin{definition}[\citet{shapley1953value}]
\label{def:shapley}
Given a training set $\dataset$ and a utility function $\U$, the Shapley value of a data point $i \in \dataset$ is defined as 
\begin{equation}
\resizebox{\columnwidth}{!}{
    $
    \begin{aligned}
    \phi_i\left(\U\right) := \frac{1}{n} \sum_{k=1}^{n} {n-1 \choose k-1}^{-1} \sum_{S \subseteq N \setminus \{i\}, |S|=k-1} \left[ \U(S \cup i) - \U(S) \right]
    \end{aligned}
    $
} \nonumber
\end{equation}
\end{definition}

The popularity of the Shapley value is attributable to the fact that it is the \emph{unique} data value notion satisfying four axioms that are usually desirable \citep{shapley1953value}. 

\textbf{Data Selection.} 
Data selection for ML is commonly formulated as an optimization problem, where the objective is to maximize the utility of the ML model based on the choice of training data. Specifically, for a given utility function $\U$, the task of \emph{size-$k$ data selection} over training set $\dataset$ is to identify the subset $\optS^{(k)}$ that optimizes:
\begin{align}
\label{eq:dataselection}
\optS^{(k)} = \argmax_{S \subseteq \dataset, |S|=k} \U(S)
\end{align}
However, solving Equation \eqref{eq:dataselection} presents significant challenges. The utility function $\U$, particularly for complex deep learning algorithms, often lacks a tractable closed-form expression for analytical optimization. A naive approach that simply evaluates the utility of all possible subsets $\U(S)$ would necessitate training ${n \choose k}$ different models, which is certainly computationally prohibitive in practical settings.  


\textbf{Data Selection via Data Shapley.} 
\add{Data selection is generally considered a standard downstream application of Data Shapley \cite{jiang2023opendataval}, where the relevant experiments can be traced back to the original Data Shapley paper \cite{ghorbani2019data}.}
The use of Data Shapley values for data selection posits that the sum $\phi(\U)[S] := \sum_{i \in S} \phi_i(\U)$ is a reliable indicator of a dataset $S$'s utility, implying a positive correlation with $\U(S)$. 
Consequently, a data selection strategy based on Data Shapley scores aims to maximize $\phi(\U)[S]$ as a proxy for optimizing $\U(S)$:
\begin{align*}
\ShapleySet^{(k)} := \argmax_{S \subseteq \dataset, |S|=k} \phi(\U)[S]
\end{align*}
Since $\phi(v)[S] = \sum_{i \in S} \phi_i$, $\ShapleySet^{(k)}$ consists of top-$k$ data points that achieve the highest Shapley values. That is, when using Data Shapley for size-$k$ data selection, \emph{the top-$k$ data points with the highest Shapley values are chosen}.\footnote{\add{We do not consider tied utilities here for simplicity, but we note that the derived results can be easily adapted to the case where multiple subsets achieve the optimal utility.}}






\section{Why Might Data Shapley Fail in Data Selection Tasks?}
\label{sec:when-fail}
The effectiveness of Data Shapley has shown mixed results. Notably, several studies \cite{wang2023threshold,kwon2023data,jiang2023opendataval} have documented that for specific datasets, model performance metrics, and selection budgets, its effectiveness can be \add{close to} random selection. This section will present a theoretical framework designed to provide insights into this puzzling phenomenon.


\subsection{A Hypothesis Testing Framework for Comparing Dataset Utilities}




\addcheck{Both the optimal data selection problem, outlined in Eqn. (\ref{eq:dataselection}), and practical data quality management tasks fundamentally involve comparing the utility of various datasets. For example, in the context of data acquisition, the focus is on determining which data source, A or B, should be chosen to augment an existing dataset \( S_0 \). This requires comparing the utility values \( v(S_0 \cup A) \) and \( v(S_0 \cup B) \). Similarly, in the case of data pruning, the goal is to identify which data points should be removed from a dataset \( S \), essentially comparing the utility of \( v(S\backslash\{i\}) \) for each element \( i \) in \( S \).
Hence, we investigate the efficacy of Data Shapley in facilitating the utility comparison for different datasets.}


Inspired by \citet{bilodeau2024impossibility}, we formulate the performance on utility comparison as a hypothesis testing problem. 
Given two subsets of training data $S_1, S_2 \subseteq \dataset$, we would like to compare their utility $\U(S_1), \U(S_2)$
without directly evaluating $\U$ on them. The null and alternative hypotheses are formulated as follows:
\begin{equation}
\begin{aligned}
&\Hnull: \U(S_1) \ge \U(S_2) \\
&\Ha: \U(S_1) < \U(S_2)
\end{aligned}
\label{eq:hypo}
\end{equation}
\textbf{Shapley value-based hypothesis test.} 
Consider a scenario where the only available information is the Shapley vector $\phi(\U) \in \R^n$. 
A \emph{Shapley value-based hypothesis test} is an arbitrary algorithm for the practitioners to draw their conclusion of the hypothesis test solely based on the Shapley vector $\phi$. 
Formally, this is a function
\begin{align*}
\testfunc: \R^n \rightarrow [0, 1]
\end{align*}
where the output of $\testfunc(\phi)$ represents the probability that the practitioner rejects $\Hnull$ (based on some external randomness). 
An example of such a test algorithm is 
$
\testfunc(\phi) = \ind \left[ \phi[S_1] < \phi[S_2] \right]
$
which is implicitly being used in Shapley value-based data selection.

\begin{remark}
In practical applications, computing Data Shapley often becomes computationally unfeasible and requires estimation through Monte Carlo methods, such as permutation sampling \citep{castro2009polynomial}. To keep the analysis clean, our study does not take the approximation error of Data Shapley into account, focusing instead on the efficacy of exact Data Shapley in data selection tasks.
\end{remark}

\begin{remark}[\textbf{All the information available to $\testfunc$ is $\phi(\U)$}]
It might be presumed that computing Data Shapley necessitates evaluating $\U(S)$ for all or a significant subset of $S$s. However, this is not always the case. For instance, the exact Data Shapley values for $K$ nearest neighbors can be efficiently calculated without the need to evaluate $\U(S)$ for any subset $S \subseteq N$ \citep{jia2019efficient, wang2023noteknn, wang2024efficient}. 
\add{Here, we assume all the information available for the hypothesis tests is the Shapley vector $\phi(\U)$ to keep the analysis clean and align with the common usage of Data Shapley for data selection.} 
\end{remark}

\subsection{Analysis}

The goal of our work is to see whether Data Shapley scores $\phi(\U)$ can reliably be used to conduct the hypothesis tests of comparing the utility of two data subsets described above. 

\textbf{Metric for evaluating hypothesis tests.} 
We adopt the classical approach of assessing hypothesis test efficacy by examining the balance between True Positive (sensitivity) and True Negative (specificity) rates, as established in the literature \cite{yerushalmy1947statistical}. 
For two datasets of interest, $S_1$ and $S_2$, we define $\classNull := \{ \U \in \R^{2^{n}-1}: \U(S_1) \ge \U(S_2) \}$ the set of all utility functions $v$ satisfying the null hypothesis, and $\classAlter := \{ \U \in \R^{2^{n}-1}: \U(S_1) < \U(S_2) \}$ the set of all utility functions $\U$ satisfying the alternative hypothesis. 
For any Shapley value-based hypothesis test $\testfunc$, the metrics are defined as:
\begin{align*}
\Spec(\testfunc) &= \inf_{v \in \classNull} \left[ 1 - \testfunc(\phi(v)) \right] \\
\Sens(\testfunc) &= \inf_{v \in \classAlter} \testfunc(\phi(v))
\end{align*}
These metrics evaluate the test's effectiveness across \emph{all} possible utility functions, with the practitioner’s goal being to maximize both True Positive and True Negative rates in terms of the hypothesis test function $\testfunc$. 
It is noteworthy that a random guessing approach, which predicts a hypothesis irrespective of Shapley values (e.g., $\testfunc(\phi) = 0.5$), achieves a combined metric of $\Spec(\testfunc) + \Sens(\testfunc) = 1$.


\textbf{Data Shapley can work no better than random guessing.}
Our analysis reveals a crucial limitation in using Shapley values for comparing dataset utilities: \emph{without specific structural assumptions about the utility functions, such tests can be no more effective than random guessing.}


\begin{theorem}
\label{thm:hypo-random}
For the utility comparison hypothesis testing problem formulated in (\ref{eq:hypo}), any Shapley value-based hypothesis test $\testfunc$ is constrained to:
\begin{align*}
\Spec(\testfunc) + \Sens(\testfunc) \le 1
\end{align*}
\end{theorem}

This theorem underscores that the maximum achievable balance between True Positive and True Negative using a Shapley value-based hypothesis test is no better than what one would expect from random guesswork. \add{Without additional information about the underlying utility function, practitioners cannot reliably predict the utility comparison between two datasets. In particular, the predictive accuracy may not surpass that of basic random guessing.}




\textbf{Underlying reasoning of Theorem \ref{thm:hypo-random}.} 
The computation of Shapley values transforms the utility function $\U$ (which can be viewed as a vector in $\R^{2^n-1}$) into the Shapley vector $\phi \in \R^n$. This transformation is not injective, allowing for the possibility that different utility functions could yield identical Shapley vectors. 
\add{Consequently, when conducting hypothesis test based on the Shapley vector $\phi(\U)$, if there exists another utility function $\U'$ (e.g., defined on a different hold-out validation set) that maps to the same Shapley vector ($\phi(\U) = \phi(\U')$) while $\U \in \classNull$ and $\U' \in \classAlter$, it becomes impossible to reliably infer the utility comparison between $S_1$ and $S_2$ based solely on the Shapley vector $\phi$.}
Hence, Theorem \ref{thm:hypo-random} immediately follows from the following:
\begin{theorem}
\label{thm:identical-Shapley}
Given any score vector $s \in \R^n$, for any dataset pair $(S_1, S_2)$, there exists two utility functions $\U$ and $\U'$ s.t. $\U \in \classNull$ and $\U' \in \classAlter$, and both yield the same Shapley vector: $s = \phi(\U) = \phi(\U')$.
\end{theorem}
The proof of the above result exploits the high-dimensional nature of the null space of the Shapley value transformation, \add{a property that is well-known in game theory but to the best of our knowledge, never has been discussed in data valuation literature.} Detailed derivation is deferred to Appendix \ref{appendix:proof}. 

\begin{remark}[Example of utility functions with identical Shapley values]
Table \ref{tb:datasets} presents a simple example where two different utility functions, $\U$ and $\U'$, result in identical Data Shapley scores. Such a situation is likely to happen in practice, where we give an analog in federated learning contexts. Imagine a validation set that is balanced and comprises data from three distinct sources $\{A, B, C\}$, and there are three clients $\{1, 2, 3\}$. In the first world, each client $1, 2, 3$ owns training data exclusively from one of $A, B, C$, leading to the utility function $\U(S) = |S|/3$. In the second world, clients ${1, 2}$ both hold data from $A$ and $B$, and client $3$ holds data from $C$. 
Since clients $1$ and $2$ holds the same training data, we have $\U'({1}) = \U'({2}) = \U'({1, 2}) = 2/3$, and $\U'({1, 3}) = \U'({2, 3}) = 1$.\footnote{We assume the data are sufficient and the model would not overfit to any of $A, B, C$, and $\U'(\{1, 2, 3\}) = 1$.} 
Despite these differences, $\U$ and $\U'$ yield the same Shapley values, $\phi(\U) = \phi(\U')$. 
Suppose we are interested in comparing the utility between $\{1, 2\}$ and $\{1, 3\}$. In the first world, they have the same utilities, while in the second world $\U'({1, 2}) < \U'({1, 3})$, which is impossible to distinguish from the Shapley values. 



\end{remark}

\begin{remark}
\add{While we state Theorem \ref{thm:hypo-random} and \ref{thm:identical-Shapley} for Data Shapley, in Appendix \ref{appendix:semivalue} we show that the result can be extended to all semivalues that satisfy the ``inverse Pascal triangle condition'' \cite{dragan2002inverse}. This includes other popular data valuation techniques such as leave-one-out error \cite{koh2017understanding} and Data Banzhaf \cite{wang2023data}.
}
\end{remark}

\begin{table}[t]
\centering
\resizebox{\columnwidth}{!}{
\begin{tabular}{ccccccccc}
\toprule
$S$    & $\varnothing$ & \textbf{\{1\}} & \textbf{\{2\}} & \textbf{\{3\}} & \textbf{\{1, 2\}} & \textbf{\{1, 3\}} & \textbf{\{2, 3\}} & \textbf{\{1, 2, 3\}} \\ \midrule
$\U$ & 0                 & 1/3            & 1/3            & 1/3            & 2/3               & 2/3               & 2/3               & 1                    \\
$\U'$ & 0                 & 2/3            & 2/3            & 1/3            & 2/3               & 1                 & 1                 & 1        \\ \bottomrule           
\end{tabular}
}
\caption{An example of two utility functions $\U, \U'$ such that $\phi_i(\U) = \phi_i(\U') = 1/3$ for all $i \in \{1, 2, 3\}$.}
\label{tb:example}
\end{table}

\section{When does Data Shapley Select Good Datasets?}
\label{sec:when-success}

\add{The preceding section shows that Data Shapley can work arbitrarily bad for data selection tasks when there are no restrictions on utility functions. However, this section will show that when the utility functions are confined to certain structures shaped by the intrinsic properties of the underlying datasets or learning algorithms, Data Shapley can be notably effective in selecting high-quality datasets.}





\subsection{\add{Illustrating Example: Heterogeneous-Quality Datasets}}
\label{sec:heterogeneous-example}

\add{We provide a simple example where Data Shapley excels: a dataset containing both high-quality and low-quality data.} In particular, we consider a dataset $\dataset = \Sgood \cup \Sbad$ comprising a mix of bad data, denoted as $\Sbad$ (such as mislabeled data or data with significant feature noise), and the remainder being high-quality, clean data, $\Sgood := \dataset \setminus \Sbad$. In this setting, Data Shapley can effectively prioritize all clean data points over the problematic ones. Specifically, for any pair of data points where $i \in \Sbad$ and $j \in \Sgood$, and for utility functions $\U$ defined by any reasonable learning algorithms, it is generally true that $\U(S \cup i) \le \U(S \cup j)$ for any subset $S \subseteq \dataset \setminus \{i, j\}$. 
\add{This is also being empirically justified in the previous literature (Figure 2 in \citet{kwon2022beta}).} 
That is, substituting any problematic data point with a clean one will not degrade machine learning model performance.


\begin{theorem}
\label{thm:structureV}
    Suppose that the dataset $\dataset$ can be divided into $\dataset = \Sbad \cup \Sgood$ where $\Sbad \cap \Sgood = \varnothing$, $|\Sgood| = k$. 
    If the utility function $\U$ fulfills the condition: 
    $
    \forall j \in \Sgood, 
    \forall i \in \Sbad, 
    \forall S \subseteq \dataset \setminus \{i, j\},~\U(S \cup i) \le \U(S\cup j)
    $
    then Data Shapley is optimal for size-$k$ data selection problem.
\end{theorem}
\add{
The proof uses an induction argument to show that $\Sgood$ is the optimal dataset for $\U$. This theorem provides insight into why Data Shapley is particularly useful for tasks such as mislabeled or noisy data detection as reported in the literature \cite{jiang2023opendataval}. 
}

\subsection{\add{A Class of ``Shapley-effective'' Utility Functions}}

\add{While Theorem \ref{thm:structureV} presents an intuitive scenario in which Data Shapley is effective in data selection, it is data-dependent and falls short of providing more insights into the structural properties of the utility functions that make them ``Shapley-effective''.
In this section, we delve deeper into identifying and describing the specific types of utility functions for which Data Shapley demonstrates effectiveness in data selection tasks.} 
Specifically, our goal is to characterize \emph{``Shapley-effective subspace''}, the set of utility functions within which Data Shapley consistently identifies the optimal subset for size-$k$ data selection problems for all $k = 1, \ldots, n-1$. It is important to note that the condition outlined in Theorem \ref{thm:structureV} only assures Data Shapley's optimality for a specific value of $k$. 


\add{Ideally, we seek to comprehensively characterize utility functions $\U$ such that $\optS^{(k)} = \ShapleySet^{(k)}$ holds true for every $k=1, \ldots, n-1$. However, developing tractable conditions that are both necessary and sufficient for the ``Shapley-effective subspace'' seems highly challenging due to the nature of the Shapley value as a weighted average across the utilities of all possible subsets. This inherent complexity limits our ability to extract any succinct conditions. Consequently, we shift our focus towards identifying sufficient conditions that can guarantee the effectiveness of Data Shapley.}

\add{
A simple yet insightful observation is that a linear function of the form $\U(S) = w_0 + \sum_{i \in S} w_i$ naturally aligns with the `Shapley-effective' criteria.\footnote{
\add{The linear utility function is being called Linear Datamodel in \citet{ilyas2022datamodels}}} 
Building on this, we consider a generalized form of linear function class, extend it through a monotonic transformation, and demonstrate that it retains the `Shapley-effective' property.}


\begin{definition}[Monotonically Transformed Modular Function (\mon)]
\label{def:Mono-Modular-Func}
A set function $\U: 2^\dataset \rightarrow \R$ is a \emph{monotonically transformed modular function} if it is of the form $\U(S) = f(w_0 + \sum_{i \in S} w_i)$, where $f: \R \rightarrow \R$ is a monotonic function, $w_0 \in \R$, and $w_i \in \R$ is the weight assigned to each $i \in N$. 
\end{definition}

\add{While the function $f$ in the definition can be either monotonically increasing or decreasing, this paper focuses exclusively on the former case. Henceforth, any reference to \mon from now on means monotonically \emph{increasing} transformed modular function.}

\begin{remark}
We note that a monotonically transformed modular function satisfies the condition in Theorem \ref{thm:structureV} if for all $i \in \Sbad$ and $j \in \Sgood$, we have $w_i \le w_j$. 
\end{remark}

\begin{theorem}
\label{thm:monosum-is-optimal}
For any utility functions $\U$ that is monotonically transformed modular, Data Shapley is optimal for size-$k$ data selection tasks for any $k = 1, \ldots, n-1$.
\end{theorem}


\mon functions are capable of capturing the utility functions of popular learning algorithms, such as kernel methods and threshold nearest neighbor classifiers. 
For instance, consider a binary classification task with a training set $\{(x_i, y_i)\}_{i=1}^n$, where the label space $y_i \in \{\pm 1\}$. 
When we use kernel method with kernel $k(\cdot, \cdot)$, the prediction $\widehat y$ on a validation point $\xval$ is given by $\widehat y = \sign(\sum_{i \in S} y_i k(x_i, \xval))$. A natural utility function for this scenario is the correctness of the prediction on the validation point $(\xval, \yval)$, where we can show that 
$
\U(S) = \ind[\yval = \widehat y] = \ind \left[ \left(\sum_{i \in S} y_i \yval k(x_i, \xval)\right) \ge 0 \right]
$. 
In this case, the utility function is a \mon function where $w_i = y_i \yval k(x_i, \xval)$ and $f(t) = \ind[t \ge 0]$.






\section{A Heuristic for Predicting Data Shapley's Optimality for General Utility Functions}
\label{sec:heuristic}


\add{
\mon represents only a specific subclass of utility functions. 
Given the diversity of utility functions encountered in practice, a natural question arises: how can we assess whether, or to what extent, a given utility function is Shapley-effective? We draw inspiration from Theorem \ref{thm:monosum-is-optimal} and propose a heuristic aimed at predicting Data Shapley's effectiveness in data selection tasks for general utility functions. 
The heuristic involves approximating the original utility function $\U$ with a \mon function $\Ualter$ that minimizes the discrepancy between $\U$ and $\Ualter$. The fitting quality of this approximation serves as a proxy for assessing the potential efficacy of Data Shapley in data selection tasks.
}




To find the best approximation to $\U$, we approach it as a supervised learning problem, where $\Ualter$ is parameterized for optimization. The ``training data'' consists of pairs of data subsets and their corresponding utility values, i.e., $\trainSample = \{(S_1, \U(S_1)), \ldots, (S_m, \U(S_m))\}$. The training objective for $\Ualter$ is to minimize the prediction error across these pairs: $\Ualter = \arg\min_{\Ualter} \sum_{j=1}^m ( \U(S_j) - \Ualter(S_j) )^2$. 
\add{After successfully fitting $\Ualter$, we assess its fitting residual $\mR_{\U}(\Ualter) := \E_{S \sim \unif(\dataset)} \left[ \left(\U(S) - \Ualter(S)\right)^2 \right]$. To account for the varying scales of different utility functions, we use \emph{normalized fitting residual} $\normR(\Ualter) := \frac{\mR_{\U}(\Ualter)}{\Var_{S \sim \unif(\dataset)}(\U(S))}$ which adjusts the fitting residual relative to the variance of the utility function, providing a more standardized measure of fit. In practice, $\normR(\Ualter)$ can be approximated using a ``validation set'' consisting of unseen data-utility pairs. A lower value of $\normR(\Ualter)$ implies that $\U$ is closely approximated by a \mon, hinting at Data Shapley's potential effectiveness.}

\begin{remark}[Computational efficiency considerations]
\add{It may initially appear that the acquisition of a training set for $\Ualter$ is computationally intensive. However, it is important to recognize that Data Shapley, frequently utilized for assessing data quality, often relies on approximation algorithms based on Monte Carlo (MC) sampling. In practical scenarios where Data Shapley scores are estimated for evaluating data quality, a substantial amount of utility samples $\{(S, \U(S))\}$ are already being generated. These samples, collected during Data Shapley's estimation process, can be effectively repurposed for fitting $\Ualter$
without necessitating additional computational overhead.\footnote{The Monte Carlo estimator for Data Shapley may have a different sampling distribution for $S$s, but we found it does not affect the fitting residual significantly.}
}
\end{remark}

\begin{remark}[High Fitting Residuals and Data Shapley's Effectiveness]
\add{Theorem \ref{thm:monosum-is-optimal} suggests that being a \mon function is a \emph{sufficient}, but not a \emph{necessary} condition, for $\U$ to be Shapley-effective. Consequently, in the cases with high fitting residuals, Data Shapley may still be effective in data selection tasks. 
Indeed, our experiments in Section \ref{sec:eval-heuristic} show that when $\normR$ is large, data selection performance tends to exhibit significant variance, making its predictability challenging. However, we observe that a moderate fitting residual still correlates strongly with data selection performance, indicating that within certain thresholds, the residual can be a reliable indicator of Data Shapley's potential efficacy.}
\end{remark}

\begin{figure*}[h]
    \centering
    \setlength\intextsep{0pt}
    \includegraphics[width=\linewidth]{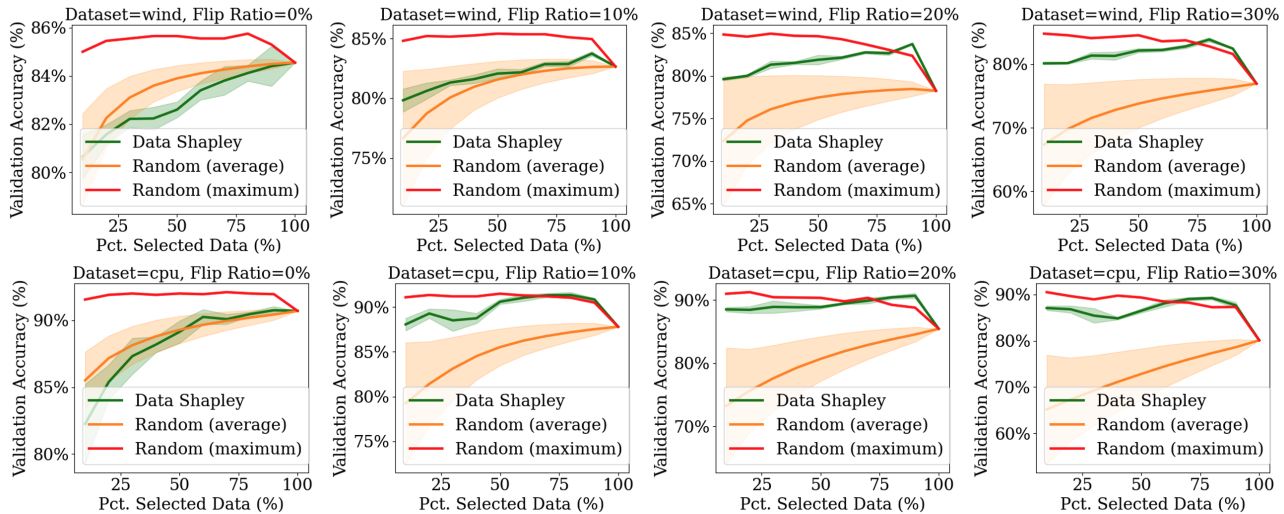}
    \caption{
    Validation accuracy curves as a function of the top $p\%$ most valuable data points added. The higher, the better. 
    `Random (average)' and `Random (maximum)' mean sample different size-$k$ subsets uniformly at random and evaluate their average and maximum utility, respectively. 
    Data Shapley's error bar indicates the standard deviation across 5 independent runs where the randomness is from the permutation sampling of Data Shapley scores. 
    }
    \label{fig:selection-main}
\end{figure*}

\subsection{When \mon function is a good approximation?} 

The utility function $\U$, being a set function determined by multiple factors such as the training set, learning algorithm, and performance metric, might be perceived as inherently complex. 
Therefore, it is interesting to understand the conditions under which $\U$ can be well-approximated by some \mon functions. 
Specifically, we explore the connection between optimal fitting residuals and the \emph{consistency index} of $\U$, an intrinsic property of utility functions. We first define the concept of $\rho$-correlated dataset pairs. 


\begin{definition}[$\rho$-correlation \cite{o2014analysis}]
\label{def:rho-correlation}
We say a pair of random variables $S, S'$ are \emph{$\rho$-correlated} if they are sampled as follows: $S$ is sampled from $\dataset$ uniformly at random ($S \sim \unif(\dataset)$), and for all $i \in S$, $i \notin S'$ w.p. $(1-\rho)/2$, and for all $i \notin S$, $i \in S'$ w.p. $(1-\rho)/2$. We use $\sampleRho(\dataset)$ to denote the distribution of $\rho$-correlated subset pairs sampled from $\dataset$. 
\end{definition}


Intuitively, commonly used learning algorithms are expected to demonstrate consistency in model behavior when trained on datasets that are similar or correlated. For example, if we have two datasets $S$ and $S'$ that are closely related (as defined by the $\rho$-correlation), the performance of models trained on these datasets should not diverge significantly as the size of the training samples increases. This expectation leads to the anticipation of a high correlation between $\U(S)$ and $\U(S')$ for utility functions that are related to test accuracy or loss. 
\add{We refer to the correlation coefficient between $\U(S)$ and $\U(S')$ for a pair of $\rho$-correlated $(S, S')$ as the \emph{\correlation} of a utility function, and we show that the high \correlation of a utility function, which suggests that minor perturbations in the training set do not lead to significant changes, serves as a positive signal for the existence of a reasonable \mon function approximation.}

\begin{theorem}
\label{thm:fit-residual-bound-main}
Let $\M$ denote the space of all \mon functions. 
For any utility functions $\U$, we have 
\begin{align*}
    \min_{\Ualter \in \M} \normR(\Ualter) 
    \le 
    \frac{1}{1-\rho^2} \left( 1 - \disc(\U) \right)
\end{align*}
where 
\begin{equation}
\resizebox{\columnwidth}{!}{
$\begin{aligned}
\disc(\U) := 
\frac{ \underset{(S, S') \sim \sampleRho(\dataset)}{\E} \left[ \U(S) \U(S') \right] - \E_{S \sim \unif(\dataset)} \left[ \U(S) \right]^2 }{ \Var_{S \sim \unif(\dataset)}(\U(S)) }
\end{aligned}$
} \nonumber
\end{equation}
is the correlation coefficient between $\U(S)$ and $\U(S')$, which we referred to as the \emph{\correlation} of $\U$.\footnote{As discussed in \citet{saunshi2022understanding}, $S$ and $S'$ have the same marginal distribution if $S \sim \unif(\dataset)$.} 
\end{theorem}


\add{The result implies that when the values between $\U(S)$ and $\U(S')$ have a stronger correlation when $S$ and $S'$ are $\rho$-correlated, the utility function $\U$ can be better approximated by \mon, which is intuitive as the mapping rules between $S$ and $\U(S)$ is more tractable.} The above theorem extends the classic result from harmonic analysis for the bound on the quality of the best linear approximation to a pseudo-boolean function in terms of its noise stability \cite{o2014analysis}. When we fix the monotonic function $f$ as the identity function $f(t) = t$, it reduces to Theorem 3.1 in \citet{saunshi2022understanding} after some rephrasing.

\begin{figure*}[h]
    \centering
    \setlength\intextsep{0pt}
    \includegraphics[width=\linewidth]{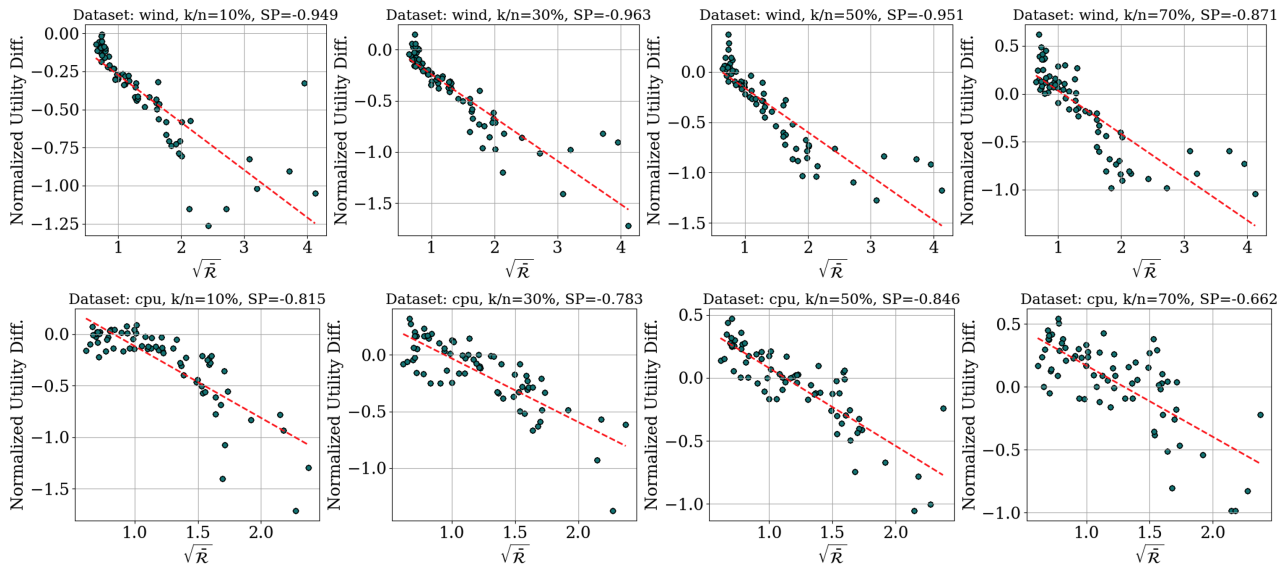}
    \caption{
    We investigate the correlation between data selection performance (measured by the normalized utility difference) and the normalized fitting residual of \mon function. 
    For each dataset, we look at size-$k$ data selection performance with $k \in \{0.1n, 0.3n, 0.5n, 0.7n\}$. 
    \add{Each point represents the results on a dataset (with different noise-flipping ratios).} 
    } 
    \label{fig:check-heuristic-main}
\end{figure*}



\section{Experiments}
\label{sec:eval}


\add{
Our experiments aim to demonstrate the following assertions: 
\textbf{(1)} Data Shapley works well when the utility functions are 
being defined on heterogeneous datasets, 
\textbf{(2)} Data Shapley's effectiveness is strongly correlated with the fitting quality of \mon functions to the utility functions, 
and \textbf{(3)} The utility functions' approximability by \mon functions is further correlated with their \correlation (deferred to Appendix \ref{appendix:eval-noise-stab}). 
In this section, to estimate Data Shapley, we use the most widely used permutation sampling estimator \cite{mitchell2022sampling}, where for each experiment the sampling budget is as high as 40,000 to reduce the instability in Shapley value estimation. 
Following \citet{ghorbani2019data,kwon2022beta}, we use logistic regression as the learning algorithm here in the main paper. 
Additional results with neural networks and detailed experiment settings are deferred to Appendix \ref{appendix:eval}.}

\subsection{When does Data Shapley work well/bad for data selection?}
\label{sec:eval-Select}

To corroborate the reasonings in Section \ref{sec:when-fail} and \ref{sec:heterogeneous-example}, we compare the efficacy of Data Shapley for data selection for datasets with different levels of varying data quality. To fairly evaluate the performance, we focus on Data Shapley's relative performance compared with the random selection baseline. Specifically, for each cardinality $k$, we sample 50,000 subsets, evaluate their utility scores, and take the average. We also show the maximum utility score among all sampled subsets. We use 40,000 utility samples for approximating Data Shapley. 


For each dataset, we create its noisy variants by randomly picking a certain proportion of the data points to flip their labels. Since mislabeled data usually negatively affect the model performance, its marginal contribution will likely be worse than any clean data points, mirroring the utility function structure described in Theorem \ref{thm:structureV}. 
As depicted in Figure \ref{fig:selection-main}, we observe that for clean datasets (i.e., Flip Ratio=0\%), the performance of subsets selected by Data Shapley marginally surpasses or worse than that of randomly chosen subsets, and significantly underperforms the subset with the maximum utility found by random selection. Conversely, in scenarios where datasets comprise data points of greater varied quality, the selection effectiveness of Data Shapley significantly improves. When the label-flipping ratio is high, Data Shapley closely matches or surpasses the highest utility found by random selection.\footnote{To better align with our discussion in Section \ref{sec:when-fail} and \ref{sec:when-success}, here the data selection performance is evaluated on the same validation set for computing Data Shapley.} 


\subsection{\mon Fitting Residual vs Data Selection Performance}
\label{sec:eval-heuristic}

\add{We evaluate the effectiveness of the heuristic we proposed in Section \ref{sec:heuristic} by assessing the correlation between \mon functions' fitting residuals and Data Shapley's performance in data selection.}


\textbf{Metric for Data Selection: normalized utility difference.} Our metric for data selection performance is the \emph{normalized utility difference}: the utility of the dataset selected by Data Shapley, $\ShapleySet^{(k)}$, compared to the optimal dataset, $\optS^{(k)}$, normalized by the utility difference between the optimal dataset and random datasets, i.e., $\frac{\U(\ShapleySet^{(k)}) - \U(\optS^{(k)})}{\U(\optS^{(k)}) - \E_{S: |S|=k}[\U(S)]}$. 
In practice, we approximate $\U(\optS^{(k)})$ and $\E_{S: |S|=k}[\U(S)]$ using the maximum and average utility of a batch of randomly sampled data subsets (same setting as in Section \ref{sec:eval-Select}). 

\textbf{Neural network implementation of \mon function.} We use a neural network-based parameterization for \mon. For a function $\Ualter(S) = f(w_0 + \sum_{i \in S} w_i)$, we encode the dataset $S$ as a binary vector $x$, where $x_i = 1$ if $i \in S$, and $x_i = 0$ otherwise. The linear combination $w_0 + \sum_{i=1}^n w_i x_i$ is implemented via a linear layer in the neural network. The monotonic function $f$ is implemented by a neural network with non-negative weight constraints. 
\add{While such an approach may not guarantee finding the optimal $\Ualter$, we empirically find that the fitting residual is fairly small (the mean squared error in most cases is $< 10^{-4}$).}


\textbf{Results.} 
\add{For each dataset, we generate 200 noisy variants, where for each of the variants we randomly flip the label of a certain portion of data points where the noise rate is uniformly sampled between 0 and 50\%. We then evaluate Data Shapley's performance in size-$k$ data selection tasks across these datasets. Meanwhile, for each of the variants, a \mon function is trained to approximate its utility function and evaluate the fitting residual $\normR$. We reuse the 40,000 utility samples initially collected for Data Shapley estimation to train the \mon model.} 
In the scatter plot in Figure \ref{fig:check-heuristic-main}, each dot corresponds to one of the dataset's variants. We present the result for data selection ratios of $k/n \in \{10\%, 30\%, 50\%, 70\%\}$. 
We can see a clear correlation between Data Shapley's performance on size-$k$ data selection task and the normalized fitting residual of the \mon function. Notably, a lower $\normR$ consistently corresponds with Data Shapley's capability to identify datasets of higher utility. 
\add{When $\normR$ is large, data selection performance indeed tends to exhibit significant variance, making its predictability challenging. This is expected as being a \mon function is only a sufficient condition for being Shapley-effective.}



\section{Conclusion \& Limitations}

This work advances the understanding of the application of Data Shapley for data selection tasks. 
We show that Data Shapley's performance can be no better than basic random selection in general settings, and we discuss the conditions under which Data Shapley excels. 


\textbf{Limitations.} 
\add{In our experiment, we demonstrate that our heuristic is highly effective when comparing Data Shapley's effectiveness among utility functions for datasets from the same domain but of different qualities. However, its applicability is less certain when comparing utility functions for datasets from different domains. This is expected as the huge differences in the function nature make their approximability or learnability not directly comparable. Despite this limitation, the heuristic remains valuable in many practical scenarios where we are dealing with datasets from the same domain but with differing qualities. In such cases, the heuristic can be used in predicting the usefulness of Data Shapley for data selection within each source.} 

\textbf{Future works.}
Building on the insights from this study, future research could explore the sufficient and necessary conditions for which Data Shapley is optimal for data selection. Additionally, a deeper exploration into the ethical implications and fairness aspects of the downstream applications of Data Shapley could be an interesting future work.


\clearpage

\section*{Acknowledgment}
This work is supported in part by the National Science Foundation under grants IIS-2312794, IIS-2313130, OAC-2239622, Amazon-Virginia Tech Initiative in Efficient and Robust Machine Learning, the Commonwealth Cyber Initiative.

We thank Tong Wu and Chong Xiang for their helpful feedback on the preliminary version of this work.

\bibliography{ref}
\bibliographystyle{icml2024}

\newpage
\appendix
\onecolumn

\section{Extended Related Works}
\label{appendix:related-work}

\subsection{Data Shapley and Friends}

\emph{Data Shapley} is one of the first principled approaches to data valuation being proposed \cite{ghorbani2019data, jia2019towards}. 
Data Shapley is based on the \emph{Shapley value}, a famous solution concept from game theory literature which is almost always being justified as the \emph{unique} value notion satisfying the following four axioms: 
\begin{enumerate}
    \item \textbf{Null player:} if $\U(S \cup \{z_i\})=\U(S)$ for all $S \subseteq D \setminus \{z_i\}$, then $\phi_{z_i}(\U)=0$. 
    \item \textbf{Symmetry:} if $\U(S \cup \{z_i\}) = \U(S \cup \{z_j\})$ for all $S \subseteq D \setminus \{z_i, z_j\}$, then $\phi_{z_i}(\U)=\phi_{z_j}(\U)$. 
    \item \textbf{Linearity:} For utility functions $\U_1, \U_2$ and any $\alpha_1, \alpha_2 \in \R$, $\phi_{z_i}(\alpha_{1} \U_{1}+\alpha_{2} \U_{2}) = \alpha_{1} \phi_{z_i}(\U_{1}) + \alpha_{2} \phi_{z_i}(\U_{2})$. 
    \item \textbf{Efficiency:} for every $\U, \sum_{z_i \in D} \phi_{z_i}(\U)=\U(D)$.
\end{enumerate}



Since its introduction, Data Shapley has rapidly gained popularity as a principled solution for data valuation. 
However, as argued in \cite{kwon2022beta}, the efficiency axiom is not necessary for the machine learning context, and the framework of \emph{semivalue} is obtained by relaxing the efficiency axiom. Moreover, \cite{lin2022measuring} provide an alternative justification for semivalue based on causal inference and randomized experiments. 
Based on the framework of semivalue, \cite{kwon2022beta} propose \emph{Beta Shapley}, which is a collection of semivalues that enjoy certain mathematical convenience. \cite{wang2023data} propose \emph{Data Banzhaf}, and show that the Banzhaf value, another famous solution concept from cooperative game theory, is the most reproducible against \emph{arbitrary} perturbation to the submodels. 
Furthermore, the leave-one-out error is also a semivalue, where the influence function \cite{koh2017understanding} is generally considered as its approximation. 
Another line of works focuses on improving the computational efficiency of Data Shapley by considering KNN as the surrogate learning algorithm for the original, potentially complicated deep learning models \cite{jia2019efficient, wang2023threshold, wang2024efficient}.  
\cite{ghorbani2020distributional, kwon2021efficient} consider Distributional Shapley, a generalization of Data Shapley to data distribution. 

\subsection{Alternative Data Valuation Methods}

There have also been approaches for data valuation that do not belong to the aforementioned types. For a detailed survey, we direct readers to \cite{sim2022data}. Notably, several studies have focused on tracking the impact of individual training examples on test loss throughout the training process \citep{pruthi2020estimating, hammoudeh2021simple, paul2021deep, yeh2022first, das2022checksel, guu2023simfluence}. Another avenue of research employs the representer theorem to decompose neural network predictions into linear combinations of training data activations \cite{yeh2009comparisons, sui2021representer}. However, \cite{sogaard2021revisiting} revealed the empirical instability of techniques such as TracIn \citep{pruthi2020estimating} and the Representer Point method \citep{yeh2018representer}.

Further, \cite{sim2020collaborative} introduced a valuation metric based on the reduction in model parameter uncertainty provided by the data. Several \emph{training-free} and \emph{task-agnostic} data valuation methods have also been proposed. For instance, \cite{xu2021validation} proposed a diversity measure known as robust volume (RV) for appraising data sources. \cite{tay2022incentivizing} devised a valuation method leveraging the maximum mean discrepancy (MMD) between the data source and the actual data distribution. \cite{nohyun2022data} introduced a \emph{complexity-gap score} for evaluating data value without training, specifically in the context of overparameterized neural networks.
\cite{wu2022davinz} applied a domain-aware generalization bound based on neural tangent kernel (NTK) theory for data valuation. \cite{amiri2022fundamentals} assessed data value by measuring statistical differences between the source data and a baseline dataset. 
\cite{just2022lava} utilized a specialized Wasserstein distance between training and validation sets as the utility function, alongside an efficient approximation of the LOO error. Lastly, \cite{kwon2023data} utilized random forests as proxy models to propose an efficient, validation-free data valuation algorithm.

\clearpage

\section{Deferred Proofs}
\label{appendix:proof}

\begin{theorem}[Restate of Theorem \ref{thm:hypo-random}]
Given any two subsets of training data $S_1, S_2 \subseteq \dataset$ such that $S_1 \ne S_2$ and $S_i \ne \varnothing$ and $S_i \ne N$ for $i \in \{1, 2\}$, the null and alternative hypothesis is formed as follows:
\begin{equation}
\begin{aligned}
&\Hnull: \U(S_1) \ge \U(S_2) \\
&\Ha: \U(S_1) < \U(S_2)
\end{aligned}
\end{equation}
Any Shapley value-based hypothesis test $\testfunc$ for the above problem is constrained to:
\begin{align*}
\Spec(\testfunc) + \Sens(\testfunc) \le 1
\end{align*}
\end{theorem}
\begin{proof}
This result immediately follows from Theorem \ref{thm:identical-Shapley}, since no matter what value score $s \in \R^n$ is available to $\testfunc$, there always exists $\U \in \classNull$ and $\U' \in \classAlter$ that yields the same Shapley vector $s = \phi(\U) = \phi(\U')$, where $\testfunc(s)$ cannot distinguish between $\U$ and $\U'$ based on $s$. 
\end{proof}

\begin{theorem}[Restate of Theorem \ref{thm:identical-Shapley}]
\label{thm:identical-Shapley-appendix}
Given any score vector $s \in \R^n$ and any two subsets of training data $S_1, S_2 \subseteq \dataset$ such that $S_1 \ne S_2$ and $S_i \ne \varnothing$ and $S_i \ne N$ for $i \in \{1, 2\}$, there exists two utility functions $\U$ and $\U'$ s.t. $\U \in \classNull$ and $\U' \in \classAlter$, and both yield the same Shapley vector: $s = \phi(\U) = \phi(\U')$.
\end{theorem}
\begin{proof}
If we view the computation of the Shapley value $\phi$ as a function from the utility function $\U$, and if we view the utility function as a size-$(2^n-1)$ vector\footnote{Recall that we assume $v(\varnothing) = 0$.}
, then the Shapley value can be viewed as a \emph{linear mapping} from $\R^{2^n-1}$ to $\R^n$. That is, $\phi = Av$ for some matrix $A \in \R^{n \times (2^n-1)}$. This can be easily inferred from the Shapley value's formula in Definition \ref{def:shapley}. 
For a given data subset $T \subseteq N$, we define a simple game $\bar u_T$ with the utility function as follows:
\begin{align*}
\bar u_T(S) = 
\begin{cases}
    1 & |S \land T| = 1 \\
    0 & \text{Otherwise}
\end{cases}
\end{align*}
Such a game is referred as \emph{commander's game} in the literature \citep{yokote2016new}, where one can show that $\phi(\bar u_T) = A \bar u_T = \zero$, and the set of $\{ \bar u_T: T \subseteq N, |T| \ge 1 \}$ forms a basis for $\R^{2^n-1}$. 
By \cite{yokote2016new}, for any utility function $\U$ with the Shapley value $s = \phi(\U)$, we can decompose it as 
\begin{align*}
v(S) = \sum_{i \in S} s_i + \sum_{T \subseteq N, |T| \ge 2} \alpha_T \bar u_T(S)
\end{align*}
Now, as long as we can show that there always exists $\{\alpha_T\}_{T \subseteq N, |T| \ge 2}$ that can form a utility function $\U$ s.t. $\U(S_1) \ge \U(S_2)$, we can construct $\U \in \classNull$ required by the theorem statement. 

\begin{align*}
\U(S_1) - \U(S_2)
&= \left( \sum_{i \in S_1} s_i + \sum_{T \subseteq N, |T| \ge 2} \alpha_T \bar u_T(S_1) \right)
- \left( \sum_{i \in S_2} s_i + \sum_{T \subseteq N, |T| \ge 2} \alpha_T \bar u_T(S_2) \right) \\
&= \sum_{i \in S_1 \setminus S_2} s_i - \sum_{i \in S_2 \setminus S_1} s_i + \sum_{T \subseteq N, |T| \ge 2} \alpha_T \left(\bar u_T(S_1) -  \bar u_T(S_2)\right)
\end{align*}

Since any of $\alpha_T$ can be set to be arbitrarily large to force $\U(S_1) - \U(S_2) \ge 0$, all we need is having $\{T: |T| \ge 2, |T \land S_1|=1, |T \land S_2| \ne 1\}$ is non-empty so that there exists at least one of $T$ s.t. $\bar u_T(S_1) -  \bar u_T(S_2) > 0$. 
This is clearly true when $S_1 \ne S_2$ and $S_i \ne \varnothing$ and $S_i \ne N$ for $i \in \{1, 2\}$. 

The construction of $\U'$ can be done similarly. 
\end{proof}

\clearpage

\begin{theorem}[Restate of Theorem \ref{thm:monosum-is-optimal}]
For any utility functions $\U$ that is monotonically transformed modular, Data Shapley is optimal for size-$k$ data selection tasks for any $k = 1, \ldots, n-1$.
\end{theorem}
\begin{proof}
Without loss of generality, let $w_1 \ge \ldots \ge w_n$. Since $f$ is a monotonic function, it is clear that the optimal size-$k$ subset $\optS^{(k)} = \{1, \ldots, k\}$. 
We now show that for such a utility function $\U$, $\phi_i(\U) \ge \phi_j(\U)$ for any $i \ge j$. 
This immediately follows from the fact that for any $S \subseteq \dataset \setminus \{i, j\}$ we have 
$
\U(S \cup i)
= f(w_0 + \sum_{\ell \in S} w_\ell + w_i) 
\ge f(w_0 + \sum_{\ell \in S} w_\ell + w_j) 
= \U(S \cup j)
$ as $w_i \ge w_j$ in the assumption. 
Since $\phi_{i} - \phi_{j}$ can be written as a positively weighted sum of $\U(S\cup i) - \U(S \cup j)$ across $S \subseteq \dataset \setminus \{i, j\}$, we have $\phi_1 \ge \phi_2 \ge \ldots \ge \phi_n$ and $\optS^{(k)}$ consists of data points with top-$k$ Data Shapley scores. 

\end{proof}

\clearpage



\begin{theorem}[Restate of Theorem \ref{thm:fit-residual-bound-main}]
Denote the subclass of monotonically transformed modular (\mon) function defined on $\dataset$ as $\M$, i.e., $\M := \{\U: \exists \text{monotonic}~f, \exists w\in \R^{n}~\text{s.t.}~\forall S \subseteq \dataset: f(w_0 + \sum_{i \in S} w_i) = \U(S)\}$.
\footnote{Recall that in this paper, `monotonic' means monotonically increasing.} 
For all $\rho \in [0, 1)$ we have 
\begin{align*}
    \min_{\Ualter \in \M} \normR(\Ualter) 
    \le 
    \frac{1}{1-\rho^2} \left( 1 - \disc(\U) \right)
\end{align*}
where 
\begin{align*}
\disc(\U) := 
\frac{ \underset{(S, S') \sim \sampleRho(\dataset)}{\E} \left[ \U(S) \U(S') \right] - \E_{S \sim \unif(\dataset)} \left[ \U(S) \right]^2 }{ \Var_{S \sim \unif(\dataset)}(\U(S)) }
\end{align*}
is the correlation coefficient between $\U(S)$ and $\U(S')$, which we referred to as the \emph{\correlation} of $\U$.\footnote{As discussed in \citet{saunshi2022understanding}, $S$ and $S'$ have the same marginal distribution if $S \sim \unif(\dataset)$.} 
\end{theorem}
\begin{proof}

Denote the monotonic function class $\F_\gamma = \{ f: \frac{\max_t f'(t)}{\min_t f'(t)} = \gamma, \min_t f'(t)>0\}$. Denote the subclass of $\M$ as $\M_\gamma = \{\U: \U(S) = f(w_0 + \sum_{i \in S} w_i), f \in \F_\gamma \}$. 

First, the space of function class $\M_\gamma$ will be remain the same if we further restrict that $f(0) = 0$, as for any $\U(S) = f(w_0 + \sum_{i \in S} w_i)$ with $f(0) \ne 0$, it can be equivalently expressed as $\U(S) = f_0\left(w_0 - f^{-1}(0) + \sum_{i \in S} w_i\right)$ with $f_0(t) = f(t + f^{-1}(0))$. 
Note that $f^{-1}(0)$ always exists due to the condition that $f'(t) \ge L$ for some constant $L > 0$. 

Now, we fix a monotonic function $f \in \F_\gamma$ s.t. $f(0) = 0$, and we denote $U := \max_t f'(t), L := \min_t f'(t)$, i.e., $\gamma = U/L$. 

Denote $g(S) = f^{-1}( \U(S) )$. From Theorem 3.1 in \citet{saunshi2022understanding}, we know that 
\begin{align*}
\min_w \E_{S} \left[ (g(S) - w_0 - \sum_{i \in S} w_i)^2 \right] 
\le \frac{1}{1-\rho^2} \left( \E_S \left[ g(S)^2 \right] - \underset{(S, S') \sim \sampleRho(\dataset)}{\E} \left[ g(S) g(S') \right] \right)
\end{align*}

Denote $w^* = \argmin_w \E_{S} \left[ (g(S) - w_0 - \sum_{i \in S} w_i)^2 \right]$. 

Since $f'(t) \in [L, U]$, we have 
\begin{align*}
\left| \U(S) - f \left(w_0 + \sum_{i \in S} w_i\right) \right| 
&= \left| f(f^{-1}( \U(S) )) - f \left(w_0 + \sum_{i \in S} w_i\right) \right| \\
&\le U \left| f^{-1}( \U(S) ) - w_0 - \sum_{i \in S} w_i \right|
\end{align*}
and since the derivative of inverse function $(f^{-1})' \in [1/U, 1/L]$, we have 
\begin{align*}
g(S) 
= f^{-1}(\U(S)) 
= f^{-1}(\U(S)) - f^{-1}(0)  
\in \left[ \frac{\U(S)}{U}, \frac{\U(S)}{L} \right]
\end{align*}

Hence, we know that 
\begin{align*}
\E_{S} \left[ \left(\U(S) - f\left(w_0^* + \sum_{i \in S} w_i^*\right)\right)^2 \right]
&\le U^2 \E_{S} \left[ \left( g(S) - w_0^* - \sum_{i \in S} w_i^* \right)^2 \right] \\
&\le \frac{U^2}{1-\rho^2} \left( \E_S \left[ g(S)^2 \right] - \underset{(S, S') \sim \sampleRho(\dataset)}{\E} \left[ g(S) g(S') \right] \right) \\
&\le \frac{U^2}{1-\rho^2} \left( \frac{1}{L^2} \E_S \left[ \U(S)^2 \right] - \frac{1}{U^2} \underset{(S, S') \sim \sampleRho(\dataset)}{\E} \left[ \U(S) \U(S') \right] \right) \\
&= \frac{1}{1-\rho^2} \left( \gamma^2 \E_S \left[ \U(S)^2 \right] - \underset{(S, S') \sim \sampleRho(\dataset)}{\E} \left[ \U(S) \U(S') \right] \right) 
\end{align*}

Therefore, we have 
\begin{align*}
    \min_{\Ualter \in \M_\gamma} \E_{S} \left[ \left(\U(S) - \Ualter(S)\right)^2 \right]
    & \le 
    \min_w \E_{S} \left[ \left(\U(S) - f\left(w_0 + \sum_{i \in S} w_i\right)\right)^2 \right] \\
    &\le 
    \frac{1}{1-\rho^2} \left( \gamma^2 \E_S \left[ \U(S)^2 \right] - 
    \underset{(S, S') \sim \sampleRho(\dataset)}{\E} \left[ \U(S) \U(S') \right] \right) \\
    &= \frac{1}{1-\rho^2} \left( \gamma^2 \left(
    \Var_S \left( \U(S) \right) + 
    \E_S \left[ \U(S) \right]^2 \right) - \underset{(S, S') \sim \sampleRho(\dataset)}{\E} \left[ \U(S) \U(S') \right] \right) \\
\end{align*}

Note that $\gamma \ge 1$. Clearly, the upper bound is minimized when $\gamma=1$. Hence, we have 
\begin{align*}
    \min_{\Ualter \in \M_\gamma} \E_{S} \left[ \left(\U(S) - \Ualter(S)\right)^2 \right]
    & \le 
    \frac{1}{1-\rho^2} \left(
    \Var_S \left( \U(S) \right) + 
    \E_S \left[ \U(S) \right]^2 - \underset{(S, S') \sim \sampleRho(\dataset)}{\E} \left[ \U(S) \U(S') \right] \right) \\
\end{align*}

Dividing both sides by $\Var_S(\U(S))$ gives the inequality in the statement. 
\end{proof}

\clearpage

\begin{theorem}[Restate of Theorem \ref{thm:structureV}]
    Suppose that the dataset $\dataset$ can be divided into $\dataset = \Sbad \cup \Sgood$ where $\Sbad \cap \Sgood = \varnothing$, $|\Sgood| = k$. 
    If the utility function $\U$ fulfills the condition: 
    $
    \forall j \in \Sgood, 
    \forall i \in \Sbad, 
    \forall S \subseteq \dataset \setminus \{i, j\},~\U(S \cup i) \le \U(S\cup j)
    $
    then Data Shapley is optimal for size-$k$ data selection problem.
\end{theorem}
\begin{proof}
We show the following two statements: 
\textbf{(1)} $\Sgood$ consists of the data points of top-$k$ Shapley values among $\dataset$ and 
\textbf{(2)} $\U(\Sgood) = \argmax_{S: S \subseteq N, |S|=k} \U(S)$. 

For \textbf{(1)}, $\Sgood$ consists of the data points of top-$k$ Shapley values among $\dataset$ since for any $j \in \Sgood$ and $i \in \Sbad$, we have $\phi_j \ge \phi_i$ which immediately follows from the condition of $\forall S \subseteq \dataset \setminus \{i, j\},~\U(S \cup i) \le \U(S\cup j)$. 
For \textbf{(2)}, we prove the following argument for any $\ell \ge 0$ with induction: 
for any $S \subseteq \dataset$ of size $k$ s.t. $|S \setminus \Sgood| = |\Sgood \setminus S| = \ell$, we have $\U(S) \le \U(\Sgood)$. 
The base case when $\ell=0$ trivially holds. 
Now, suppose that the statement holds true for all $\ell \le L-1$. Now, consider an $S$ where $|S \setminus \Sgood| = L$. 
Consider an alternative $S' \subseteq \dataset$ and $|S'|=k$ where 
$S' \setminus S = \{j\}$ for some $j \in \Sgood$ and $S \setminus S' = \{i\}$ for some $i \in \Sbad$. 
That is, $S'$ is constructed by removing an $i \in \Sbad$ by an $j \in \Sgood$. 
Let $S'' := S \cup S'$. We have $\U(S') = \U(S'' \cup j) \ge \U(S'' \cup i) = \U(S)$. 
Moreover, since $|S' \setminus \Sgood| = L=1$, by induction hypothesis we have $\U(S') \le \U(\Sgood)$, which implies that $\U(S) \le \U(\Sgood)$. 
\end{proof}

\clearpage

\subsection{Extension of Theorem \ref{thm:identical-Shapley} to Semivalue}
\label{appendix:semivalue}

\emph{Semivalue} \cite{dubey1981value} is originally studied in cooperative game theory. It has recently been proposed as a unified framework for data value notions \citep{kwon2022beta, lin2022measuring} which comprises many existing data value notions such as LOO \cite{koh2017understanding}, Data Shapley \cite{ghorbani2019data}, Beta Shapley \cite{kwon2022beta}, and Data Banzhaf \cite{wang2023data}.  
The popularity of semivalues is attributable to the fact that they are the collection of all possible data value notions that satisfy three important axioms: dummy player, symmetry, and linearity. 
The specific definition of the three axioms can be found in Appendix \ref{appendix:related-work}. 

\begin{definition}[Semivalues]
\label{def:semi-value}
We say a data value notion is a semivalue if and only if it satisfies the linearity, dummy player, and symmetry axioms. 
\end{definition}

The following theorem shows that every semivalue of a data point $i$ can be expressed as a weighted average of marginal contributions $\U(S \cup i)-\U(S)$ across different subsets $S \subseteq \dataset \setminus i$. 

\begin{theorem}[Representation of Semivalue \citep{dubey1981value}]
\label{thm:semivalue}
A value function $\phi$ is a semivalue, if and only if, there exists a set of weights $\{\alpha^{(n)}_k, k=1, \ldots, n\}$ such that $\sum_{k=1}^{n} {n-1 \choose k-1} \alpha^{(n)}_k = 1$ and the value function $\phi$ can be expressed as follows:
\begin{equation}
\begin{aligned}
\phi_{i}\left( \U \right) \nonumber := \frac{1}{n} \sum_{k=1}^{n} \alpha^{(n)}_k
\sum_{ \substack{S \subseteq \dataset \setminus  i,\\ |S|=k-1}}
\left(\U(S \cup i) - \U(S)\right)
\end{aligned}
\end{equation}
\end{theorem}

For example, when $\alpha^{(n)}_k = \frac{1}{n} {n-1 \choose k-1}^{-1}$, it reduces to the Shapley value. 
When $\alpha^{(n)}_k = \frac{1}{2^{n-1}}$, it reduces to the Banzhaf value. 
When $\alpha^{(n)}_k = \ind[k = n]$, it reduces to the LOO error. 

\begin{definition}[inverse Pascal triangle condition \cite{dragan2002inverse}]
We say a semivalue $\phi$ with weights coefficients $\alpha^{(n)}_k$ satisfies the ``inverse Pascal triangle condition'' if 
    \begin{align*}
        \forall t = 1, 2, \ldots, \forall k \in \{1,2,\cdots, t-1\}: \alpha_{k}^{(t-1)} = \alpha_{k}^{(t)} + \alpha_{k+1}^{(t)}
    \end{align*}
\end{definition}

We can easily verify that this condition is satisfied for all of LOO, the Shapley value, and the Banzhaf value. 

\begin{theorem}
    \label{thm:identical-Shapley-extension}
    Given a semivalue $\phi$ with weights coefficients $\alpha^{(n)}_k$, if it satisfies the ``inverse Pascal triangle condition'', then for any score vector $s \in \R^n$ and any two subsets of training data $S_1, S_2 \subseteq \dataset$ such that $S_1 \ne S_2$ and $S_i \ne \varnothing$ and $S_i \ne N$ for $i \in \{1, 2\}$, there exists two utility functions $\U$ and $\U'$ s.t. $\U \in \classNull$ and $\U' \in \classAlter$, and both yield the same semivalue vector: $s = \phi(\U) = \phi(\U')$.    
\end{theorem}
\begin{proof}
Similar to the proof for Theorem \ref{thm:identical-Shapley}, we exploit the null space of semivalue. 

For a given data subset $T \subseteq N$, we define a utility function $w_T$ as follows:
$$
w_T(S) = \left\{\begin{aligned}
        & \sum_{c=0}^{|S| - |T|} {(-1)^c\left(\begin{matrix}|S| - |T|\\ c\end{matrix}\right)}/\alpha_{c+|T|}^{c+|T|} & S\supsetneq T\\
        & 1/\alpha_{|T|}^{|T|} & S = T\\
        & 0 & S \subsetneq T
    \end{aligned}\right.
$$

By \cite{dragan2002inverse}, for any semivalue $\phi$ that satisfies the inverse Pascal triangle condition, $\{ w_T: T \subseteq \dataset, T \ne \emptyset \}$ is a basis for the space of utility functions, and for any utility function $\U$ with the semivalue $s = \phi(\U)$, we can decompose it as 
\begin{align*}
\U(S) = 
\sum_{|T| \le n - 2} \beta_{T} w_T(S) + \beta_{\dataset} \left(w_\dataset(S) + \sum_{i\in \dataset} w_{\dataset \setminus i}(S)\right) - \sum_{i \in \dataset} s_i w_{\dataset \setminus i}(S)
\end{align*}

Moreover, by \cite{dragan2002inverse}, the set $\{w_T(S): 1 \le |T| \le n - 2 \} \cup \{w_\dataset + \sum_{i\in \dataset} w_{\dataset \setminus i}\}$ is a basis for the null space of semivalue $\phi$, i.e., $\phi(w) = 0$ for all utility functions $w$ from this set. 

Now, as long as we can show that there always exists $\{\beta_T\}_{T \subseteq N}$ that can form a utility function $\U$ s.t. $\U(S_1) \ge \U(S_2)$, we can construct $\U \in \classNull$ required by the theorem statement. 

$$
    \begin{aligned}
    v(S_1) - v(S_2) 
        = &\sum_{|T| \le |N| - 2} \beta_{T} (w_T(S_1) - w_T(S_2)) \\
        & +\beta_\dataset \left(w_\dataset(S_1) - w_\dataset(S_2) + \sum_{i \in \dataset} [w_{\dataset \setminus i}(S_1) - w_{\dataset \setminus i}(S_2)]\right) \\
        & - \sum_{i\in \dataset} s_i\cdot(w_{\dataset \setminus i}(S_1) - w_{\dataset \setminus i}(S_2))
    \end{aligned}
$$
Since $\beta_{T}(v)$ can set to be arbitrarily large to make $v(S_1) - v(S_2) \ge 0$, all we need is having at least one $w_T(S_1) \ne w_T(S_2)$, which is clearly true when $S_1\ne S_2$ and none of $S_1$ or $S_2$ equals $\dataset$ or $\varnothing$ . 
    
The construction of $v'$ can be done similarly using previous techniques.

\end{proof}

\clearpage

\section{Additional Settings \& Experiments}
\label{appendix:eval}

\subsection{Datasets \& Architectures}
\label{appendix:datasets}

\paragraph{Datasets.}
An overview of the dataset information we used in Section \ref{sec:eval} can be found in Table \ref{tb:datasets}. 
These are commonly used datasets in the existing literature in data valuation \cite{ghorbani2019data, kwon2022beta, jia2019towards, wang2023data, kwon2023data, wang2024efficient}. 
Following \citet{kwon2022beta}, for the datasets that have multi-class, we binarize the label by considering $\ind[y=1]$. 
Given the large amount of model retraining required in our experiment, for each of the dataset we take a size-200 subset as the training set, and a size-2000 subset as the validation set. 
This is the same as prior studies in Data Shapley \cite{kwon2022beta, wang2023data}. 

\begin{table}[h]
\centering
\begin{tabular}{@{}cccc@{}}
\toprule
\textbf{Dataset} &  \textbf{Source}                        \\ \midrule
Wind              &\url{https://www.openml.org/d/847}   \\
CPU               &\url{https://www.openml.org/d/761}   \\
Fraud             &\cite{dal2015calibrating}            \\
2DPlanes          &\url{https://www.openml.org/d/727}    \\
Vehicle           &\cite{duarte2004vehicle}             \\
Apsfail           &\url{https://www.openml.org/d/41138} \\
Pol               &\url{https://www.openml.org/d/722}   \\ \bottomrule
\end{tabular}
\caption{A summary of datasets used in Section \ref{sec:eval}'s experiments.}
\label{tb:datasets}
\end{table}


\paragraph{Architectures.}
In the experiments in the main paper, we use logistic regression as the learning algorithm. Here in Appendix, we also show the results when using a two-layer MLP model as the learning algorithm, where there are 100 neurons in the hidden layer, activation function ReLU, batch size $128$, (initial) learning rate $10^{-2}$ and Adam optimizer for training. 

\paragraph{Architecture for training \mon function.} 
In Section \ref{sec:eval-heuristic} and Appendix \ref{appendix:eval-noise-stab}, we use a neural network-based parameterization for \mon. For a function $\Ualter(S) = f(w_0 + \sum_{i \in S} w_i)$, we encode the dataset $S$ as a binary vector $x$, where $x_i = 1$ if $i \in S$, and $x_i = 0$ otherwise. The linear combination $w_0 + \sum_{i=1}^n w_i x_i$ is implemented via a linear layer in the neural network. 
The monotonic function $f$ is implemented by a neural network with non-negative weight constraints. \add{While such an approach may not guarantee finding the optimal $\Ualter$, we find that the fitting residual is fairly small (the mean squared error in most cases is $< 10^{-4}$).} 
We use an MLP with 2 hidden layers to implement the monotonic function $f$, where each layer has 100 neurons. We add an attention layer between the first and the second hidden layer. 
We reuse the 40,000 utility samples collected from Shapley value estimation for the training and testing of \mon, where we split the utility samples into 32,000 for training and 8,000 for evaluation. 
We use batch size $32$, (initial) learning rate $10^{-3}$, and Adam optimizer for training 10 epochs.

\subsection{Additional Experiments for Section \ref{sec:eval-Select}}
\label{appendix:eval-dataselection}

\begin{figure*}[htbp]
    \centering
    \includegraphics[width=\linewidth]{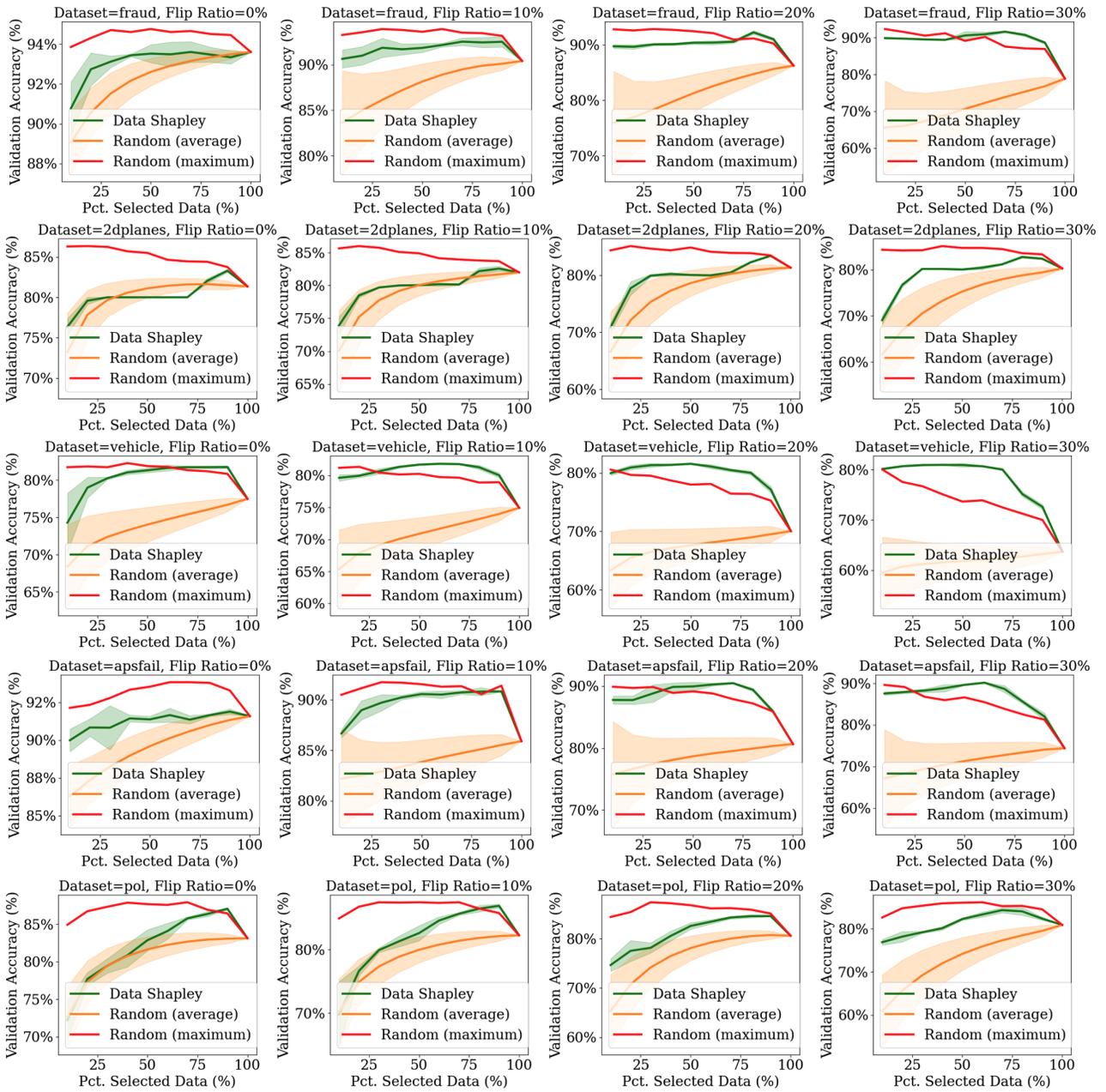}
    \caption{
    Additional results when using logistic regression classifiers. 
    Validation accuracy curves as a function of the most valuable data points added. 
    The higher, the better. 
    `Random (average)' and `Random (maximum)' means sample different size-$k$ subsets uniformly and random and evaluate their average and maximum utility, respectively. 
    Data Shapley's error bar indicates the standard deviation across 5 independent runs where the randomness is from the permutation sampling of Data Shapley scores. }
    \label{fig:selection-appendix-one}
\end{figure*}

\begin{figure*}[htbp]
    \centering
    \includegraphics[width=\linewidth]{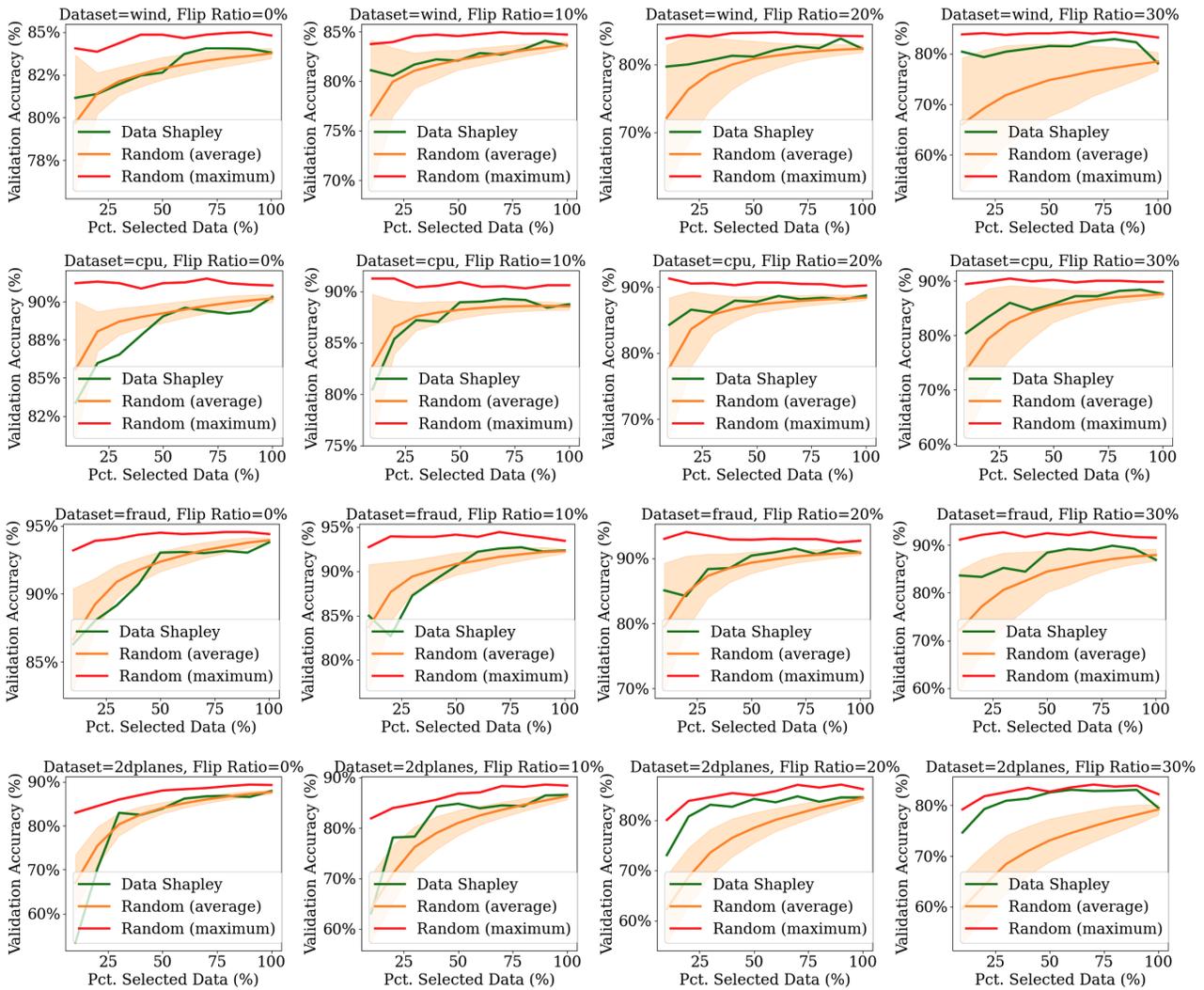}
    \caption{    
    Additional results when using MLP classifiers. 
    The figure shows the validation accuracy curves as a function of the most valuable data points added. The higher, the better. 
    `Random (average)' and `Random (maximum)' means sample different size-$k$ subsets uniformly and random and evaluate their average and maximum utility, respectively. 
    Data Shapley's error bar indicates the standard deviation across 5 independent runs where the randomness is from the permutation sampling of Data Shapley scores. }
    \label{fig:selection-appendix-MLP}
\end{figure*}

In Figure \ref{fig:selection-appendix-one}, we show the data selection results for additional datasets when using logistic regression classifiers. 
In Figure \ref{fig:selection-appendix-MLP}, we show the data selection results when using MLP classifiers. We note that in this case, because there is randomness during model training, the maximum utility found by random selection baseline can be higher than other approaches when trained on full datasets. 
The results are similar to what is observed from the maintext: for clean datasets, Data Shapley's performance is much worse than that of their noisy variants, which corroborates the insights that without additional constraints on the underlying utility function's characteristics (such as \add{the quality of particular data points}), the performance of Data Shapley can be no better than random guessing.

\clearpage

\subsection{Additional Experiments for Section \ref{sec:eval-heuristic}}
\label{appendix:eval-heuristic}

\begin{figure*}[h]
    \centering
    \setlength\intextsep{0pt}
    \includegraphics[width=\linewidth]{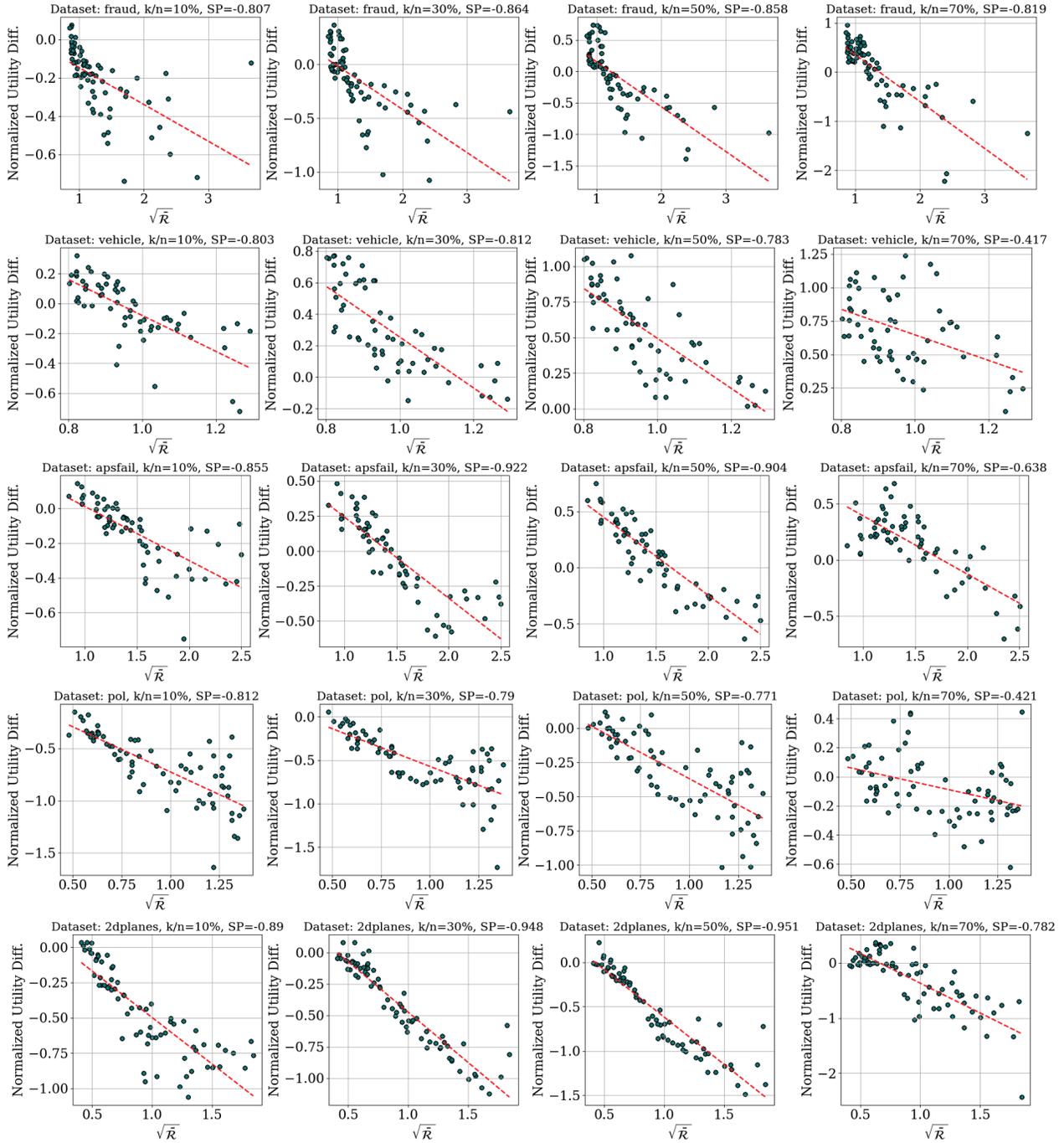}
    \caption{
    Results on additional datasets for the correlation between $\normR$ and data selection performance. 
    We investigate the correlation between data selection performance and the normalized fitting residual of \mon function. For each dataset, we look at size-$k$ data selection performance with $k \in \{0.1n, 0.3n, 0.5n, 0.7n\}$. 
    \add{Each point represents the results on a dataset (with different noise-flipping ratios).} 
    } 
    \label{fig:check-heuristic-appen}
\end{figure*}

\begin{figure*}[h]
    \centering
    \setlength\intextsep{0pt}
    \includegraphics[width=\linewidth]{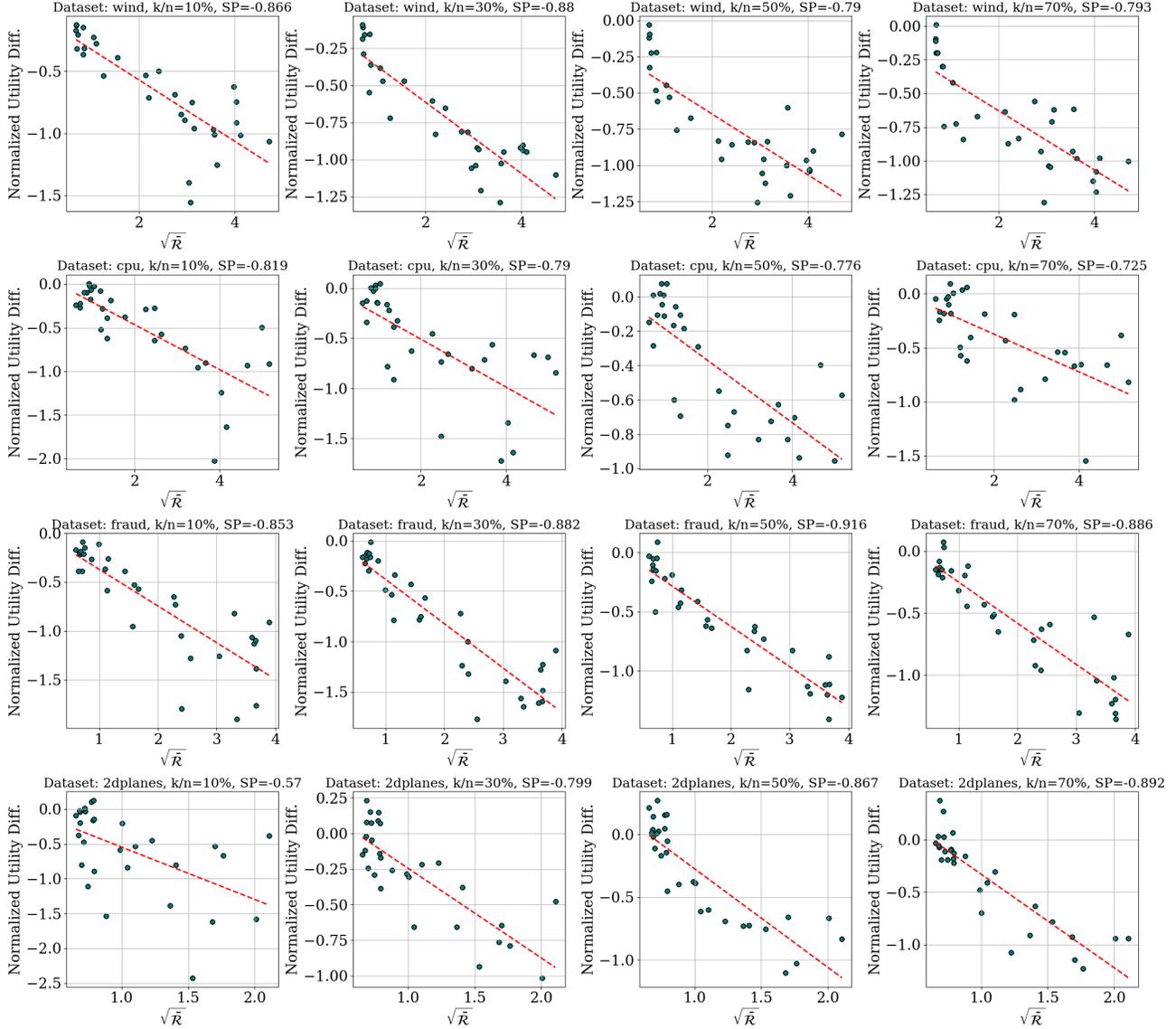}
    \caption{
    Results for the correlation between $\normR$ and data selection performance when using MLP classifier. 
    We investigate the correlation between data selection performance and the normalized fitting residual of \mon function. For each dataset, we look at size-$k$ data selection performance with $k \in \{0.1n, 0.3n, 0.5n, 0.7n\}$. 
    \add{Each point represents the results on a dataset (with different noise-flipping ratios).} 
    } 
    \label{fig:check-heuristic-MLP}
\end{figure*}

In Figure \ref{fig:check-heuristic-appen}, we show the results for comparing Data Shapley's data selection performance and $\normR$ on additional datasets, and in Figure \ref{fig:check-heuristic-MLP} we show additional results on MLP classifiers. 
Similar to the results in the maintext, the fitting residual of \mon function and Data Shapley's performance exhibit a strong correlation, which further validates the effectiveness of the heuristic proposed in Section \ref{sec:heuristic}.

\clearpage

\subsection{\mon Fitting Residual vs $\rho$-Consistency Index}
\label{appendix:eval-noise-stab}


In this experiment, we investigate the correlation between the fitting residual of \mon function and \correlation $\disc(\U)$ defined in Theorem \ref{thm:fit-residual-bound-main}. The setting in this experiment is the same as the one in Section \ref{sec:eval-heuristic}, and we additionally compute the \correlation for each noisy variant. 
Following the theorem's guidance, our focus is on scenarios with relatively low noise rates ($\rho \in \{0, 0.1, 0.2, 0.3\}$). 
For each specified value of $\rho$, we generate 5000 pairs of $\rho$-correlated datasets $S, S'$ and estimate \correlation $\disc(\U)$. 

\begin{figure*}[h]
    \centering
    \setlength\intextsep{0pt}
    \includegraphics[width=0.8\linewidth]{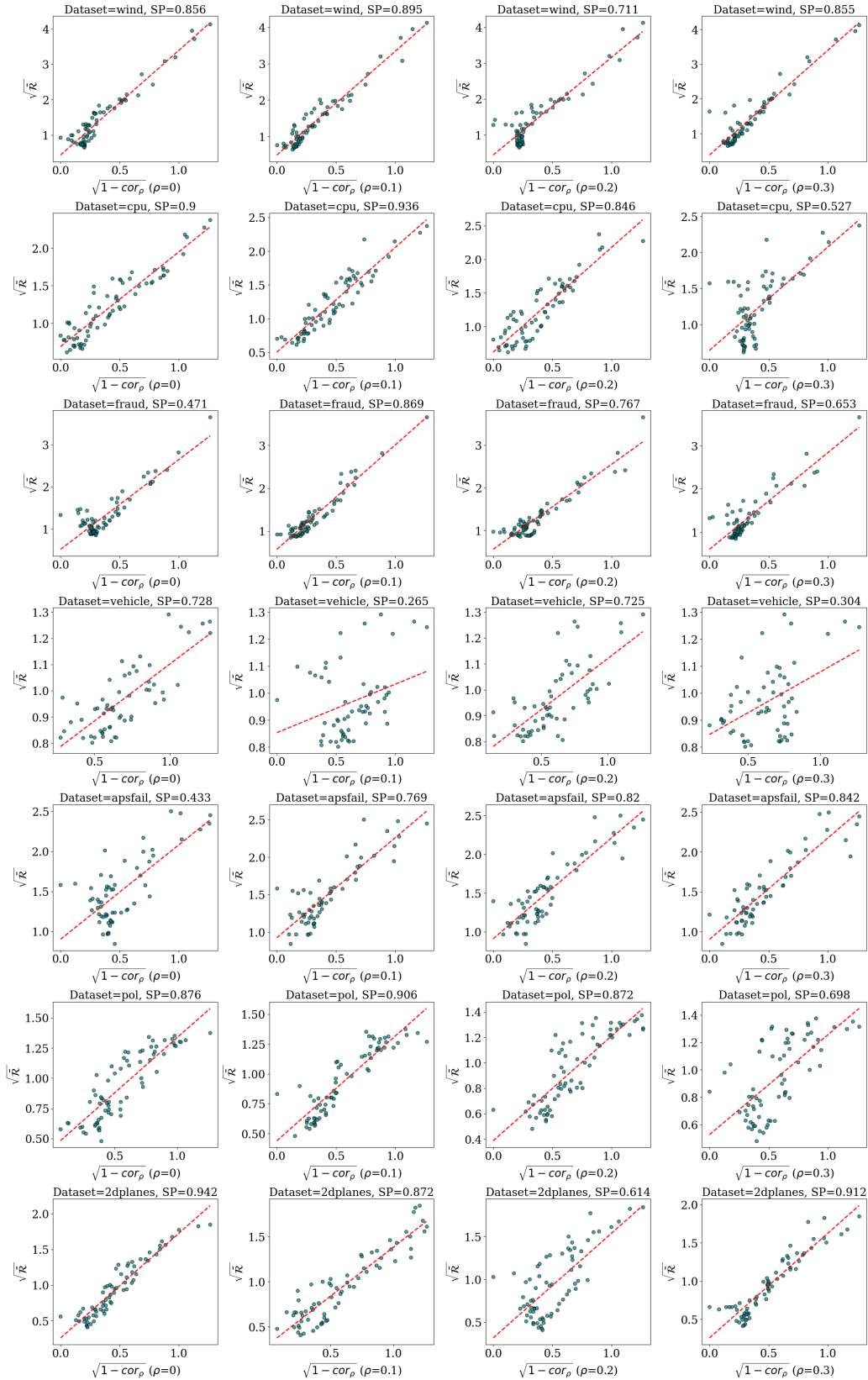}
    \caption{We investigate the correlation between the normalized fitting residual of \mon and the \correlation of the utility functions. The results for other values of $\rho$s are deferred to Appendix \ref{appendix:eval}. 
    We vary different $\rho \in \{0, 0.1, 0.2, 0.3\}$ and estimate the \correlation. 
    } 
    \label{fig:residual-vs-NS-appen}
\end{figure*}

\begin{figure*}[h]
    \centering
    \setlength\intextsep{0pt}
    \includegraphics[width=\linewidth]{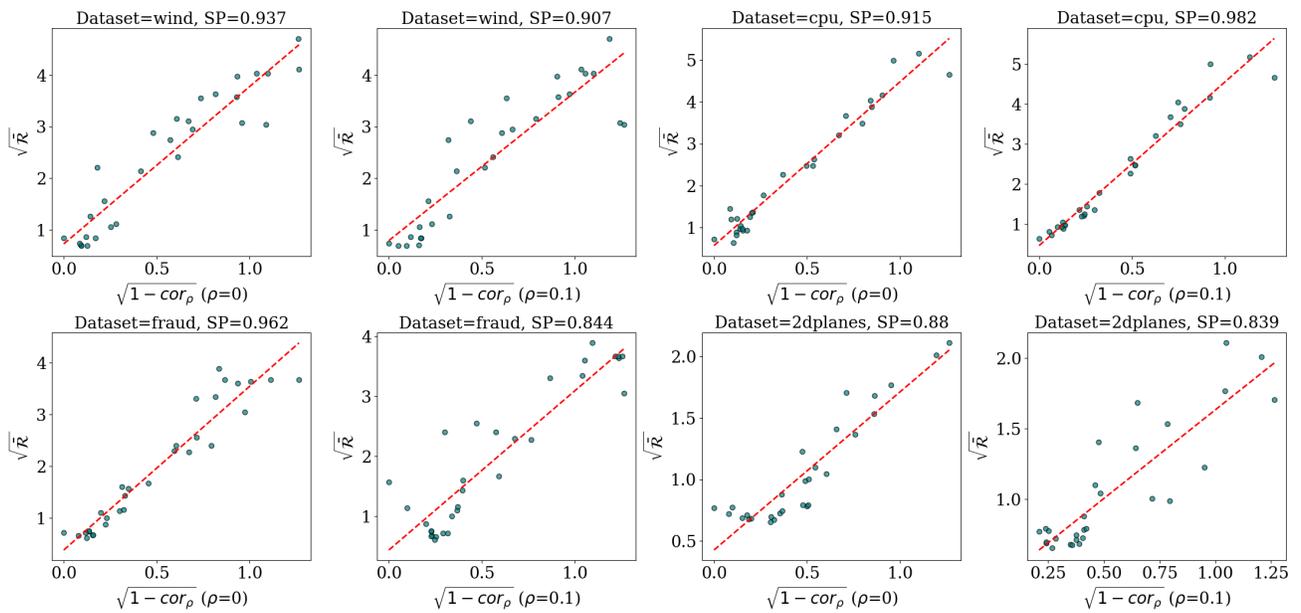}
    \caption{We investigate the correlation between the normalized fitting residual of \mon and the \correlation of the utility functions. The results for other values of $\rho$s are deferred to Appendix \ref{appendix:eval}. We vary different $\rho \in \{0, 0.1\}$ and estimate the \correlation. 
    } 
    \label{fig:residual-vs-NS-MLP}
\end{figure*}

\begin{figure*}[h]
    \centering
    \setlength\intextsep{0pt}
    \includegraphics[width=0.8\linewidth]{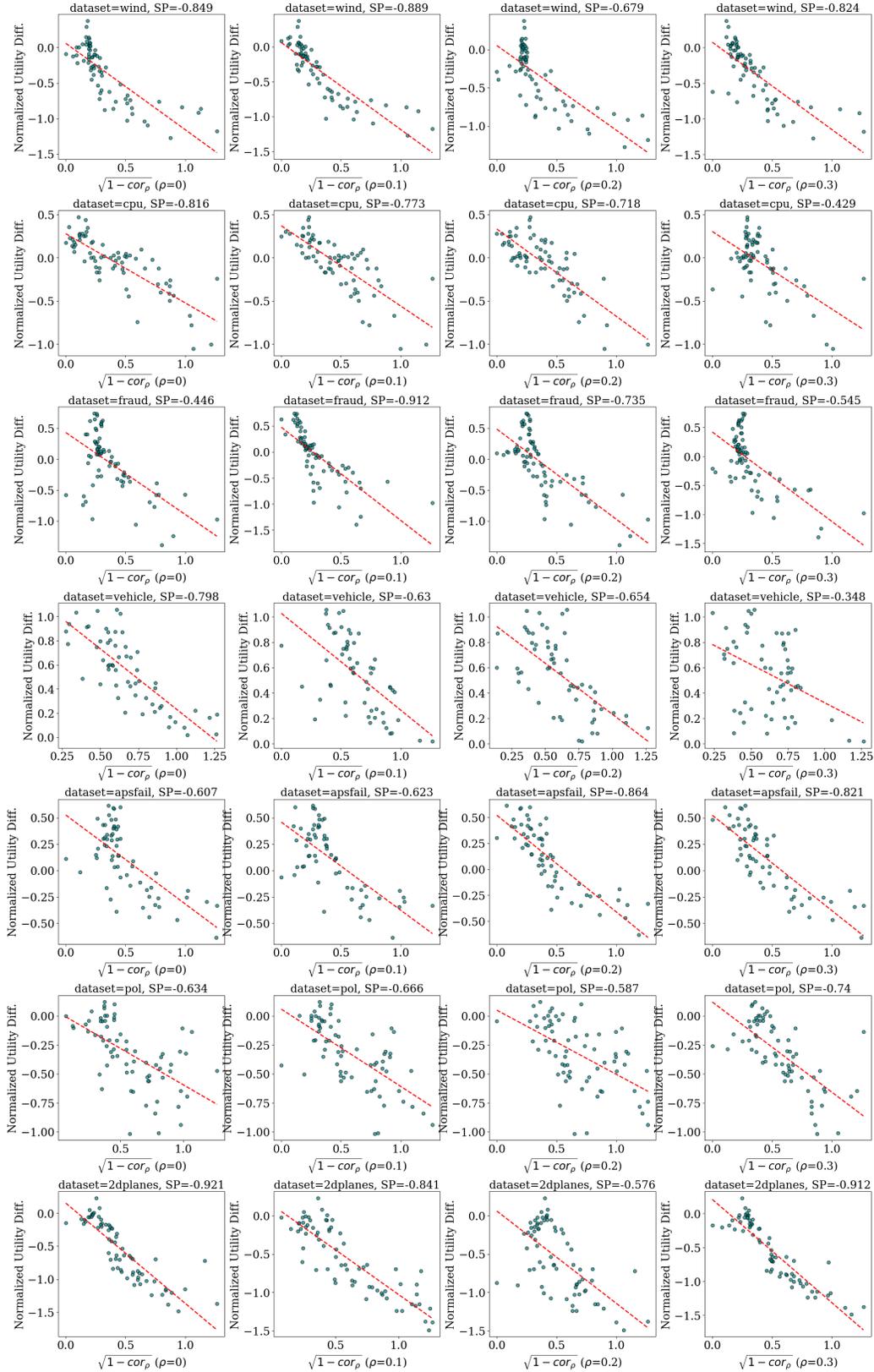}
    \caption{We investigate the correlation between the normalized fitting residual of \mon and the \correlation of the utility functions. 
    The results for other values of $\rho$s are deferred to Appendix \ref{appendix:eval}. 
    We vary different $\rho \in \{0, 0.1, 0.2, 0.3\}$ and estimate the \correlation. 
    } 
    \label{fig:select-vs-NS-appen}
\end{figure*}

\begin{figure*}[h]
    \centering
    \setlength\intextsep{0pt}
    \includegraphics[width=\linewidth]{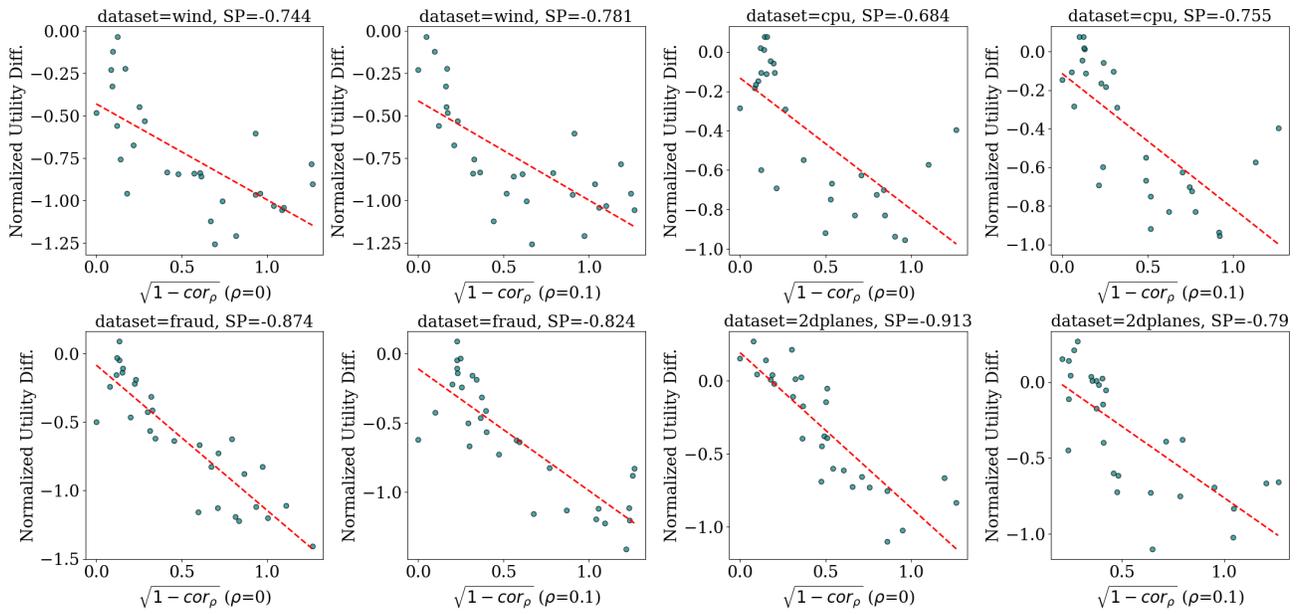}
    \caption{We investigate the correlation between the normalized fitting residual of \mon and the \correlation of the utility functions. 
    The results for other values of $\rho$s are deferred to Appendix \ref{appendix:eval}. 
    We vary different $\rho \in \{0, 0.1\}$ and estimate the \correlation. 
    } 
    \label{fig:select-vs-NS-MLP}
\end{figure*}

Figure \ref{fig:residual-vs-NS-appen} and \ref{fig:residual-vs-NS-MLP} show the results on logistic and MLP classifier, respectively. 
As we can see, there is a strong correlation between \correlation and the fitting residual $\normR(\U)$. This observation lends empirical support to the theoretical assertions made in Theorem \ref{thm:fit-residual-bound-main}, suggesting that \correlation is indeed a significant factor in determining the fitting quality of \mon functions to utility functions. Since \mon's fitting residual is correlated to the Data Shapley's data selection performance as we have shown earlier, $\disc(\U)$ is also highly correlated with Data Shapley's data selection performance, as validated in Figure \ref{fig:select-vs-NS-appen} and \ref{fig:select-vs-NS-MLP}. 
This is because for noisy datasets, since two correlated subsets are likely to both contain similar amounts of bad data, their utilities have a stronger correlation compared with clean datasets where data points are of similar quality. 


\end{document}